\documentclass[review,12pt,numbers]{elsarticle}
\usepackage{amsmath,booktabs,graphicx,hyperref,natbib,array,xcolor,longtable,caption}
\graphicspath{{figs/}}
\usepackage[margin=2.5cm]{geometry}
\usepackage{pdfpages}

\journal{To be decided}

\begin{document}

\begin{frontmatter}

\title{Electricity price forecasting across
Norway's five bidding zones in the post-crisis era}

\author[1]{My Thi Diem Phan\corref{cor1}}
\author[2]{Tuyen Trung Truong}
\author[3]{Hoai Phuong Ha}
\author[4]{Dat Thanh Nguyen\corref{cor1}}
\address[1]{Independent researcher}
\address[2]{Department of Mathematics, University of Oslo}
\address[3]{Department of Computer Science, The Arctic University of Norway}
\address[4]{Faculty of Medicine, University of Oslo}

\cortext[cor1]{Corresponding author emails are diemmyphan@outlook.com and thanhdng@uio.no.}

\begin{abstract}
Norway's electricity market is heavily dominated by hydropower, but the 2021-2022 energy crisis and stronger integration with Continental Europe have fundamentally altered price formation, reducing the reliability of forecasting models calibrated on historical data. Despite the critical need for updated models, a unified benchmark evaluating feature contributions across all structurally diverse Norwegian bidding zones remains lacking. Here we present a comprehensive evaluation of one-step-ahead forecasting of the Nord Pool market across all five Norwegian bidding zones. We constructed a multimodal hourly dataset spanning 2019–2025 and evaluated eight forecasting model families, including Light Gradient Boosting Machine (LightGBM), autoregressive models with exogenous variables, and advanced deep learning architectures, using a strictly causal test set. We implemented robust rolling-origin backtesting, leave-one-group-out feature ablation, and conditional regime analysis to dissect model performance and feature utility. Our results show that LightGBM achieves the best performance in every zone, with mean absolute error ranging from 1.60 to 5.58 euros per megawatt-hour, while a ridge-regularized autoregressive model with exogenous variables remains a highly competitive linear benchmark in northern zones. Feature ablation reveals that models relying solely on lagged prices and calendar variables achieve high accuracy and often match or closely approach the performance of the full multimodal model. However, conditional regime analysis demonstrates that external features like reservoir levels and gas prices remain crucial to stratify forecast errors, which consistently increase under stressed market regimes. This highlights the practical value of model interpretability and regime awareness for decision makers facing structural changes in market dynamics.

\end{abstract}


\end{frontmatter}


\section{Introduction}
\label{sec:intro}

Electricity price forecasting (EPF) became important after the liberalisation of European electricity markets in the late 1990s \cite{EUDirective1996}. In day-ahead markets, prices respond quickly to short term supply and demand and can move sharply within the same day \cite{weron2014}. For many years, this task was relatively stable, and autoregressive models captured much of the predictable daily and weekly structure \citep{weron2014,lago2021}. That setting changed in 2021 and 2022. Low gas storage, post-pandemic demand recovery, and the disruption of Russian gas supply pushed title transfer facility (TTF) gas prices to extreme levels and changed electricity price formation across Europe \cite{ESMA2023TTF,saether2024effect,ACER2022MMR,ACER2022Assessment}. Models trained on pre-crisis data therefore became less reliable in the new regime.

Norway is a useful setting for studying this problem. Around 88\% of Norwegian electricity production comes from hydropower \cite{NVE2024}, but the market is split into five bidding zones NO1-NO5, with clear structural differences. The southern zones are more exposed to Continental European price signals, while the northern zones remain more hydro dominated and usually have lower price levels. This contrast became stronger after NordLink and North Sea Link entered full operation in 2021 and expanded the transmission of external price pressure into southern Norway \cite{Statnett2021nsl,Statnett2021nordlink}. Norway therefore offers a strong test case for post-crisis electricity price forecasting across zones with different market dynamics.

The broader EPF literature is extensive, but it does not fully answer this Norwegian post-crisis question. Foundational reviews showed that no single class of model dominates across markets \citep{weron2014}. Later work established the importance of regularised linear models, careful feature design, and long out of sample evaluation \citep{uniejewski2019,lago2021,marcjasz2020}. Recent reviews also documented the move toward machine learning and deep learning methods \citep{jedrzejewski2022}. However, much of the benchmark evidence still comes from pre-crisis data or from markets other than Norway \citep{lago2021}.

For Norway and the Nordic market, the evidence is still limited. Earlier studies focused on the Nord Pool system price, weekly settings, or cross border effects before the structural break \citep{kristiansen2012,kristiansen2014,li2021}. Other work considered monthly Nordic aggregate forecasting, short crisis-period windows, or settings without strict causal evaluation \citep{mehrdoust2023,vamathevan2022eem,omdena2025}. A recent study on NO4 examined a long term scenario setting rather than a short-horizon benchmark \citep{kth2025}. As a result, the literature still lacks a unified post-crisis benchmark across all five Norwegian bidding zones for hourly one-step-ahead forecasting of the Nord Pool day-ahead price, together with rigorous evaluation and analysis of feature contributions.

\begin{table}[!htbp]
\caption{Summary of EPF studies on the Norwegian and Nordic electricity
market prior to the present work.
Artificial neural network (ANN); autoregressive model with exogenous inputs (ARX);
Diebold-Mariano test (DM); Extreme Gradient Boosting (XGBoost);
genetic algorithm (GA); leave-one-group-out feature ablation (LOGO);
Light Gradient Boosting Machine (LightGBM); long short-term memory network (LSTM);
long-term scenario forecast (LT); seasonal autoregressive integrated moving average with exogenous inputs (SARIMAX);
temporal convolutional network (TCN); $\bullet$\,=\,yes;\;$\circ$\,=\,partial;\;$\times$\,=\,no.}
\label{tab:nordic_comparison}
\centering
\footnotesize
\setlength{\tabcolsep}{4pt}
\renewcommand{\arraystretch}{1.25}
\begin{tabular}{%
  >{\raggedright\arraybackslash}p{2.8cm}  
  >{\centering\arraybackslash}p{1.7cm}    
  >{\centering\arraybackslash}p{1.4cm}    
  >{\raggedright\arraybackslash}p{3.0cm}  
  >{\centering\arraybackslash}p{2.0cm}    
  >{\centering\arraybackslash}p{1.1cm}    
  >{\centering\arraybackslash}p{1.0cm}    
  >{\centering\arraybackslash}p{1.1cm}    
}
\toprule
\textbf{Study} &
\textbf{Coverage} &
\textbf{Resol.} &
\textbf{Method(s)} &
\textbf{Test period} &
\textbf{Causal split} &
\textbf{DM test} &
\textbf{LOGO} \\
\midrule

Kristiansen~\cite{kristiansen2012} &
  System price & Hourly &
  Autoregressive (AR) &
  Pre-2012 &
  $\times$ & $\times$ & $\times$ \\[2pt]

Kristiansen~\cite{kristiansen2014} &
  System price & Weekly &
  Linear reg.\ + hydro vars &
  Pre-2010 &
  $\times$ & $\times$ & $\times$ \\[2pt]

Li and Becker~\cite{li2021} &
  System price & Hourly &
  LSTM + market coupling &
  Pre-2021 &
  $\times$ & $\times$ & $\times$ \\[2pt]

Mehroudust et al.~\cite{mehrdoust2023} &
  Nordic aggr. & Monthly &
  ANN + genetic algorithm &
  Pre-crisis &
  $\times$ & $\times$ & $\times$ \\[2pt]

Vamathevan et al.~\cite{vamathevan2022eem} &
  All 5 zones & Hourly &
  ANN, LSTM &
  Onset 2021 &
  $\times$ & $\times$ & $\times$ \\[2pt]


Henje Scott and W. Mellander~\cite{kth2025} &
  NO4 only & Annual (LT) &
  Linear + XGBoost &
  2024 (LT) &
  $\times$ & $\times$ & $\times$ \\[2pt]

Omdena~\cite{omdena2025}$^{\,\dagger}$ &
  NO1-NO4 & Hourly &
  Ridge, SARIMAX &
  In-crisis 2022 &
  $\times$ & $\times$ & $\times$ \\

\midrule

\textbf{Our study} &
  \textbf{All 5 zones} & \textbf{Hourly} &
  \textbf{LightGBM, XGBoost, Ridge ARX, LSTM, TCN, Transformer, Naïve-24h/168h} &
  \textbf{Full 2025} &
  $\bullet$ & $\bullet$ & $\bullet$ \\

\bottomrule
\multicolumn{8}{l}{%
  \footnotesize $^{\dagger}$Non-peer-reviewed technical blog report.}\\
\end{tabular}
\end{table}

Table~\ref{tab:nordic_comparison} summarises the main Norwegian and Nordic studies most relevant to the present work which makes the gap clear. No earlier study combined all five Norwegian zones, a full post-crisis test year, strict causal evaluation, Diebold-Mariano testing, and leave one group out feature ablation in one framework. This study addresses that gap, we build a multimodal hourly dataset for 2019-2025 and evaluate eight model families on a strictly causal 2025 test set. Our main contributions are fourfold.

\begin{enumerate}

\item \textbf{Multimodal dataset.} We construct a public hourly dataset for all five Norwegian bidding zones from 2019 to 2025 by combining European Network of Transmission System Operators for Electricity (ENTSO-E) power system data, Norwegian Water Resources and Energy Directorate (NVE) reservoir statistics, Open-Meteo weather reanalysis, and Yahoo Finance commodity prices.

\item \textbf{Post-crisis benchmark.} We evaluate two na\"{i}ve baselines, a ridge autoregressive model with exogenous inputs (ARX), Light Gradient Boosting Machine (LightGBM), Extreme Gradient Boosting (XGBoost), and three deep learning models on a common next-hour forecasting task under a strictly causal split with training on 2019-2023, validation on 2024, and testing on 2025. We also include a 52-step rolling-origin backtest and Diebold-Mariano significance tests.

\item \textbf{Pre-crisis versus post-crisis data.} We show that data from 2019 to 2021 alone do not capture the post-crisis regime well, but they still improve performance when combined with post-crisis training data.

\item \textbf{Feature and regime analysis.} We quantify the contribution of six feature groups through leave one group out ablation and examine forecast errors across reservoir anomaly and TTF gas price regimes.

\end{enumerate}

\section{Materials and methods}

\subsection{Data sources}
\label{sec:data}
We construct an hourly panel dataset covering five Nord Pool bidding
zones (NO1-NO5) from 1~January 2019 to 30~December 2025, yielding 61,338
rows per zone (306,690 rows in the combined panel). Four independent sources are integrated as shown in Supplementary Table~S1.

The target variable is the day-ahead electricity price in EUR/MWh
per bidding zone. Norwegian prices exhibit pronounced positive
skewness showed in Supplementary Table~S2 (skewness 2.9-5.0 across zones), heavy tails (kurtosis 12-54), and frequent negative price episodes (0.7-1.1\% of hours),
particularly in hydro-surplus periods. The energy crisis of 2021-2022
produced mean annual prices of 192.5 EUR/MWh in NO1 and 211.3
EUR/MWh in NO2, compared to a 2025 mean of 58.3 and 65.3 EUR/MWh
respectively, illustrating the structural regime shifts that challenge
any fixed-parameter forecasting model.

\begin{figure}[!htb]
  \centering
  \includegraphics[width=\linewidth]{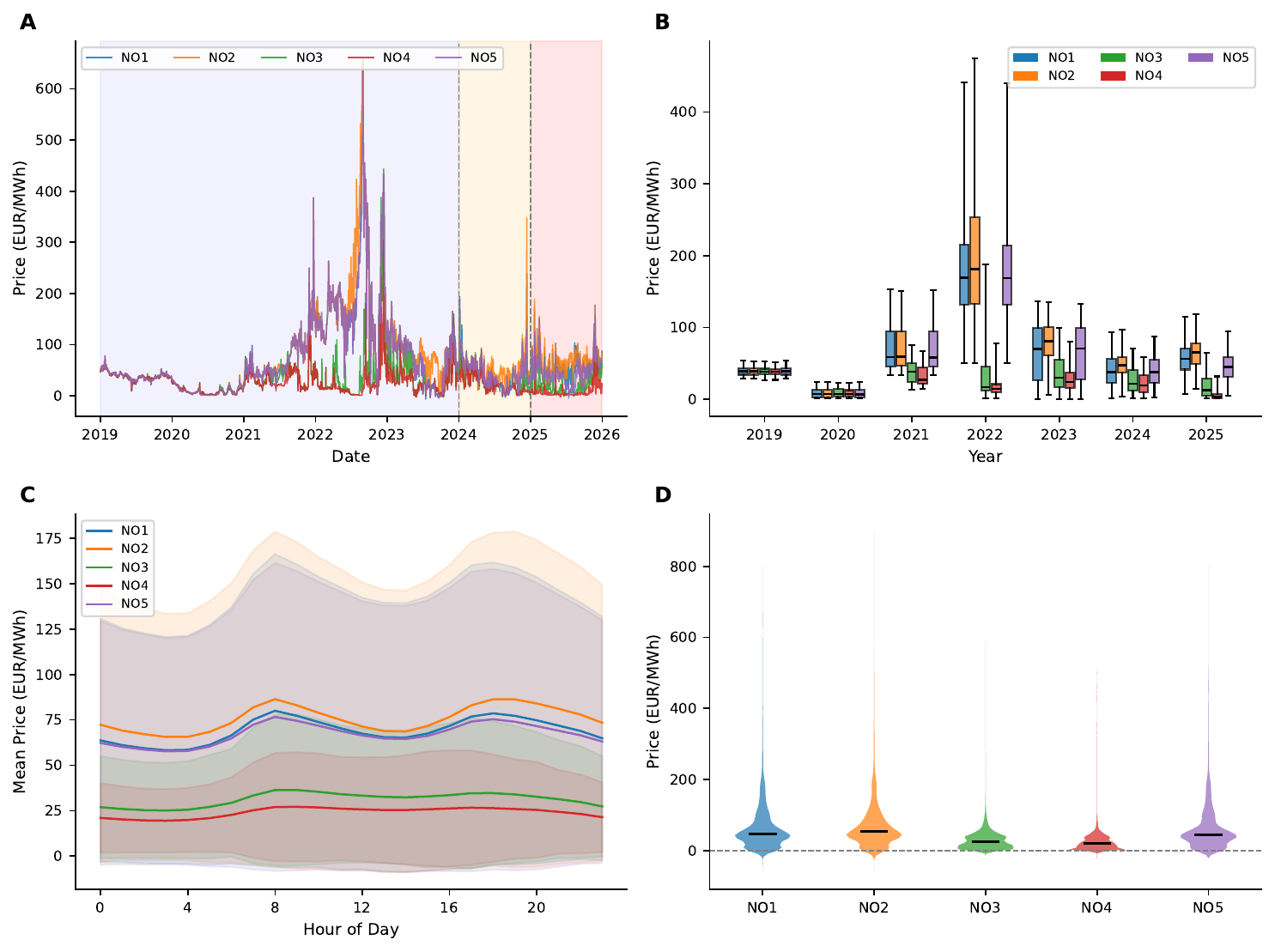}
  \caption{Price overview across all five Norwegian bidding zones
    (NO1-NO5), 2019-2025.
    \textbf{(A)}~Daily mean prices coloured by zone;
    \textbf{(B)}~Annual price distributions (box plots, interquartile range [IQR] shading);
    \textbf{(C)}~Hourly price profiles by zone;
    \textbf{(D)}~Violin distributions of hourly prices by zone.
    The 2021-2022 European energy crisis produces extreme values
    (NO$_1$ max $\approx 800$~EUR/MWh). Dashed horizontal lines mark zero.}
  \label{fig:price_overview}
\end{figure}

Figure~\ref{fig:price_overview} provides a full characterisation of 
the target variable across all five bidding zones and the entire 
2019-2025 study period. Panel~A shows daily mean prices coloured by 
zone, with the training period (2019-2023), validation year (2024), 
and test year (2025) marked by background shading. The 2021-2022 
European energy crisis is immediately visible as a sustained price 
surge across all zones, with annual mean prices in NO1 and NO2 
reaching 192.5 and 211.3~EUR/MWh respectively in 2022, compared to 
9.3 and 9.3~EUR/MWh in 2020. Panel~B summarises the annual price 
distributions as box plots and confirms the structural regime shift.
The spread of prices in 2022 far exceeds the full price range observed 
across 2019-2020 for every zone. Panel~C plots the mean hourly price 
profile averaged over the full period, revealing a pronounced morning 
peak centred on hour 08:00 and an evening shoulder between 17:00 and 
20:00, driven by residential and industrial load cycles. Panel~D presents violin 
distributions of hourly prices by zone. The southern zones NO1, NO2, 
and NO5 show higher absolute price levels and wider absolute spreads, 
with full-period means of 69.0, 75.7, and 67.1~EUR/MWh respectively, 
while the northern zones NO3 and NO4 are more compact in absolute 
terms, with means of 31.2 and 24.1~EUR/MWh. All five zones exhibit 
positive skewness and heavy upper tails, with kurtosis values ranging 
from 12.4 in NO1 to 54.3 in NO4.

\subsection{Data preprocessing}
All sources are merged onto a master hourly coordinated universal time (UTC) index derived from
ENTSO-E day-ahead prices. Weekly reservoir data are forward-filled to
hourly resolution, so each hour inherits the most recently
published weekly value. Daily commodity prices are similarly
forward-filled to hourly resolution. All timestamps are normalised to
Europe/Oslo with correct daylight saving time handling, using central European time (CET) and central European summer time (CEST).
Columns with $>$50\% missing values are dropped (e.g., generation
solar in NO1-NO2). Remaining edge-period gaps are forward-then-back
filled. No imputation is applied to the lag warm-up period (first 168
rows), which is excluded from modelling. A crucial leakage-prevention step excludes all post-delivery
variables. Four column groups are removed as post-delivery actuals:
actual generation by type (\texttt{gen\_*}), actual realised load
(\texttt{load\_mw}), actual cross-border physical flows
(\texttt{flow\_*}), and actual net imports
(\texttt{net\_import\_*}). These are only observable after the
delivery hour and would constitute look-ahead leakage. Only the
ENTSO-E-published load forecast and wind/solar forecast available
before the target hour are retained. Cross-zone correlations are high between the southern zones,
where $r > 0.95$, but much lower between north and south,
where $r = 0.31$-$0.46$, motivating zone-specific
models rather than a single pooled model.

\subsection{Feature engineering}
\label{subsec:features}
Features are organised into six groups as detailed in Supplementary Table~S3. Calendar features use sine cosine projections to preserve circular topology  (e.g., hour 23 and hour 0 are adjacent in encoding space). All lagged-price features (historical electricity prices observed 1, 2, 3, 6, 12 and other hours before the prediction time) use causal shifts causal shifts, with rolling statistics shifted by one period prior to windowing to exclude the current hour. This prevents any information from the delivery hour or later from entering the feature set, ensuring strict causal integrity throughout the feature matrix. More broadly, the feature construction is designed to reflect the one-step-ahead information set. At time $t-1$, the model predicts $p_t$ using only quantities observed or published by $t-1$, avoiding any look-ahead bias. Weekly reservoir data are forward filled to hourly resolution, such that each hour inherits the most recently published weekly release. Daily commodity prices are similarly forward filled to hourly resolution, ensuring consistency across data sources. Seasonality characteristics, including monthly means, autocorrelation function (ACF), day of week profiles, and temperature price scatter, are visualised in Supplementary Figure~S1.


\subsection{Temporal split and evaluation protocol}
\label{sec:split}

We follow the fixed-origin evaluation protocol recommended by 
Lago et al. \cite{lago2021}, which requires a held-out test set of at least 
364 days that post-dates all training and model selection data. 
The training set spans 1 January 2019 to 31 December 2023 
(43,824 hours), the validation set covers the full year 2024 
(8,784 hours), and the test set runs from 1 January to 
30 December 2025 (8,736 hours). The 2021-2022 European energy crisis, characterised by Russian 
gas supply curtailment and record TTF gas and European Union Allowance (EUA) carbon prices, represents 
a structural break in price formation. Using 2024 as the validation 
year ensures that hyperparameter selection is performed on 
post-crisis data, while keeping the full 2025 test set unseen 
during all model development decisions. Gradient boosting models use the validation set for early stopping 
(patience 50 rounds). Ridge regression selects the regularisation 
parameter $\lambda \in \{0.1, 1.0, 10.0, 100.0\}$ using the 2024 
validation set as the held-out fold. Additionally, we conduct a 52-step weekly rolling-origin backtest 
across all five zones (260 steps total) to provide a distribution 
of forecast errors rather than a single point estimate 
\citep{marcjasz2020}. 

\subsection{Models}
\subsubsection{Na\"{i}ve baselines}

Two na\"{i}ve baselines are included as reference benchmarks 
following standard practice in the EPF 
literature \citep{weron2014}. Na\"{i}ve-24h sets $\hat{p}_t = 
p_{t-24}$, repeating the price observed at the same hour on the 
preceding day, and reflects the day-ahead market structure in 
which prices exhibit strong hour-of-day periodicity. Na\"{i}ve-168h 
sets $\hat{p}_t = p_{t-168}$, repeating the price from the same 
hour one week prior, capturing the additional weekly seasonality 
driven by workday and weekend demand cycles. Any proposed model 
that fails to outperform both baselines provides no practical 
value over a simple repetition rule. In the post-crisis Norwegian 
market, both baselines produce substantially higher errors than 
in the pre-crisis period, reflecting the increased price 
volatility introduced by continental gas price coupling.

\subsubsection{Linear autoregressive model with exogenous inputs and ridge regularisation}

The linear baseline is an autoregressive model with exogenous 
inputs (ARX) of the form

\begin{equation}
    \hat{p}_t = \mathbf{x}_t^\top \mathbf{w} + b,
\end{equation}

where $\mathbf{x}_t$ contains the features and $\mathbf{w}$ is estimated by minimising 
the $\ell_2$-penalised least-squares objective

\begin{equation}
    \mathcal{L}(\mathbf{w}) = \|\mathbf{p} - X\mathbf{w}\|_2^2 
    + \lambda \|\mathbf{w}\|_2^2,
\end{equation}

with regularisation strength $\lambda \in \{0.1, 1.0, 10.0, 
100.0\}$ selected on the 2024 validation set.  
All features are standardised to zero mean and unit variance prior 
to fitting; predictions are inverse-transformed to the original 
scale. Ridge regularisation is particularly appropriate here 
because the input matrix contains highly correlated lag and 
commodity features, where ordinary least squares would be 
numerically unstable. This model serves as the primary linear 
baseline against which all non-linear methods are compared.

\subsubsection{Light gradient boosting machine}

LightGBM is a gradient boosted decision tree framework that grows trees leaf wise rather than level wise, allowing more flexible tree structures and faster reduction in training loss for a given number of leaves \citep{ke2017lightgbm}. It is trained on the full feature set described in Supplementary Table~S3 using mean absolute error (MAE) as 
the training objective, which is more robust to the extreme 
price spikes that characterise the post-crisis Norwegian market 
than the standard mean squared error loss.

The model is configured with 63 leaves per tree, a learning 
rate of 0.05, row and column subsampling rates of 0.8, and 
$\ell_2$ regularisation strength of 1.0. These values were 
selected based on validation set performance. A maximum of 
1,000 boosting rounds is used, with early stopping triggered 
after 50 consecutive rounds without improvement on the 2024 
validation MAE to prevent overfitting.

\subsubsection{Deep learning architectures}

Three deep learning architectures are benchmarked. These are a long short-term memory network (LSTM) \citep{hochreiter1997long}, a temporal convolutional network (TCN) \citep{bai2018empirical}, and a 
transformer encoder \citep{vaswani2017transformer}. All three models share a common training configuration with a 168-hour context window 
corresponding to one full week of hourly prices and a direct 
24-step multi-output head. During training, the network predicts the next 24 hourly prices jointly using MAE as the 
training loss, the AdamW optimiser with linear warm-up and 
cosine annealing over a maximum of 150 epochs with early 
stopping patience of 20 epochs. For comparability with the 
linear and tree-based benchmarks, all reported validation and 
test metrics are computed from the first output step only, 
i.e. the next-hour forecast. All input features are 
standardised to zero mean and unit variance; predictions are 
inverse-transformed to the original price scale. Each model 
also includes a linear shortcut that maps the most recent 
time-step directly to the output, initialising the network 
close to a linear ARX solution and accelerating convergence.

The LSTM consists of six stacked recurrent layers with a 
hidden dimension of 128, dropout of 0.15, and layer 
normalisation applied to the input sequence. The TCN uses 
six dilated causal residual blocks with dilation factors 
$\{1, 2, 4, 8, 16, 32\}$, a kernel size of 3, and a model 
dimension of 128, yielding a receptive field of 253 hours 
that fully covers the 168-hour context window. Group 
normalisation and dropout of 0.1 are applied within each 
block. The Transformer uses six pre-layer-normalisation 
encoder layers with model dimension 128, four attention 
heads, a feed-forward dimension of 256, sinusoidal positional 
encoding, and dropout of 0.1. All three models are of 
comparable parameter count (612k-813k per zone), ruling out 
model capacity as an explanation for any observed performance 
differences. Each model is trained independently for each of 
the five bidding zones.

\subsection{Evaluation metrics}

Model accuracy is assessed using four complementary metrics 
following standard practice in EPF
\citep{weron2014}. MAE is the primary ranking metric,

\begin{equation}
    \text{MAE} = \frac{1}{T} \sum_{t=1}^{T} |p_t - \hat{p}_t|,
\end{equation}

chosen because it is interpretable in the original price unit 
(EUR/MWh), robust to the occasional extreme price spikes that 
inflate squared-error metrics, and consistent with the MAE 
training objective used by LightGBM. Root mean squared error is reported as a complementary metric that penalises 
large errors more heavily,

\begin{equation}
    \text{RMSE} = \sqrt{\frac{1}{T} \sum_{t=1}^{T} 
    (p_t - \hat{p}_t)^2}.
\end{equation}

The symmetric mean absolute percentage error (sMAPE) is included as a scale-free relative 
accuracy measure \citep{makridakis1993},

\begin{equation}
    \text{sMAPE} = \frac{1}{T} \sum_{t=1}^{T} 
    \frac{2|p_t - \hat{p}_t|}{|p_t| + |\hat{p}_t|} \cdot 100,
\end{equation}

bounded between 0 and 200 per cent regardless of price sign 
or magnitude. Standard mean absolute percentage error (MAPE) is not used because Norwegian 
prices include near-zero and negative values that inflate the 
denominator. The coefficient of determination $R^2$ is 
reported as a variance-explained measure, where a value of 
1.0 indicates a perfect fit and values below zero indicate 
that the model performs worse than a constant mean predictor, 
as observed for the Naïve-168h baseline in several zones.

\subsection{Statistical testing}

Pairwise forecast accuracy is assessed using the Diebold–Mariano (DM) test \citep{diebold1995}, which evaluates the null hypothesis $H_0$ that two competing models have equal expected predictive loss against the one-sided alternative $H_1$ that model A achieves lower mean squared error than model B. All tests are one-sided, as the objective is to evaluate whether more complex models improve upon simpler benchmark specifications.

We apply the Harvey-Leybourne-Newbold (HLN) small-sample correction \citep{harvey1997}, which adjusts the test statistic for finite-sample bias and evaluates it against a Student $t$-distribution rather than the standard normal distribution. The variance is estimated using a Newey–West heteroscedasticity and autocorrelation consistent (HAC) estimator with bandwidth 23 lags as a conservative correction for autocorrelation in hourly loss differentials. All tests are conducted on the full test set.

Statistical significance is reported using standard thresholds. These are $p < 0.05$ (*), $p < 0.01$ (**), and $p < 0.001$ (***), with ns denoting not significant. For each of the five zones, we report three pairwise comparisons, LightGBM vs. Ridge ARX, LightGBM vs. Naïve-24h, and Ridge ARX vs. Naïve-24h.

\subsection{Pre-crisis versus post-crisis data}

To assess how the training window affects post-crisis forecasting, we conduct an additional experiment with LightGBM. We compare three training windows while keeping the validation and test periods fixed. The full-window variant is trained on 2019-2023. The post-crisis variant is trained on 2022-2023. The pre-crisis variant is trained on 2019-2021. All three variants use the same feature set, the same 2024 validation year for model selection, and the same 2025 test set for final evaluation. This design isolates whether pre-crisis observations remain useful when forecasting performance is assessed entirely on post-crisis data.

\subsection{Feature group ablation}

\label{sec:ablation}

A key methodological question in EPF is whether
Norway-specific exogenous variables, such as reservoir anomaly statistics and
commodity prices, provide useful predictive information beyond the dominant
autoregressive signal. To evaluate this directly, we adopt a LOGO ablation design instead of standard tree-based importance scores, which
can be unreliable under strong feature correlation and do not directly reflect
out-of-sample forecast utility. This approach measures each feature group's
marginal contribution to test performance by comparing the full model with a
model trained after removing that group.

The ablation is performed with LightGBM for all five bidding zones, as this
model achieves the lowest test MAE in every zone (see section~\ref{sec:results}).
For each zone, the number of boosting rounds is fixed to the value selected by
early stopping in the corresponding full model, so that differences in
performance reflect feature removal rather than re-tuning effects. The six
feature groups considered are lags, calendar, weather, reservoir, commodities,
and \texttt{load\_wsf}. In the LOGO experiment, each feature group is removed in
turn and the model is retrained on the remaining variables. The effect of
removing a group is measured as $\Delta \mathrm{MAE} =
\mathrm{MAE}_{\text{without group}} - \mathrm{MAE}_{\text{full}}$, where a
positive value indicates that the group improves forecast accuracy, while a
negative value suggests that the group is redundant or slightly noisy in the
presence of the others. We also evaluate the corresponding change in explained
variance using $\Delta R^2 = R^2_{\text{full}} - R^2_{\text{without group}}$,
so that positive values again indicate a useful contribution from the removed
group.

To complement the LOGO analysis, we conduct group-only experiments on top of
the autoregressive baseline. In these runs, the lag group is always retained,
and each exogenous group is added one at a time. Performance is summarized as
$\Delta \mathrm{MAE} = \mathrm{MAE}_{\text{lags+group}} -
\mathrm{MAE}_{\text{full}}$ and $\Delta R^2 = R^2_{\text{lags+group}} -
R^2_{\text{full}}$, which quantify how closely each reduced model approaches
the full model when combined only with lagged prices. Together, these two
analyses distinguish between a group's unique marginal value in the full model
and its additive value beyond the lag baseline. This design provides a more
interpretable assessment of feature utility than conventional importance
scores, particularly when predictors are strongly correlated.

\subsection{Conditional regime analysis}
\label{sec:regime}

To examine how forecast accuracy changes under different market conditions, we
perform a conditional regime analysis using two state variables available at
forecast time. These are reservoir anomaly and the TTF natural gas price. Reservoir
anomaly represents the hydro state relative to its seasonal normal, whereas TTF
gas price captures external thermal-market pressure transmitted to the Nordic
market. For each bidding zone, the 2025 test set is partitioned into four regime cells using a binary split for each variable. Reservoir conditions are classified as
high or low according to the median reservoir anomaly in the test set, and the
TTF regime is defined analogously using the test-set median gas price. This
yields four combinations, high-reservoir/high-TTF, high-reservoir/low-TTF,
low-reservoir/high-TTF, and low-reservoir/low-TTF. Median splits are used to
ensure balanced regime cells and stable error estimates.

The analysis is carried out for two models. These are the full LightGBM model and the
reduced Lags+Calendar specification. For each regime cell, forecast performance
is evaluated using sMAPE and RMSE. This design makes it possible to assess not
only whether prediction errors increase under stressed hydro or gas conditions,
but also whether the full feature set provides a meaningful advantage over the
reduced autoregressive-seasonal model in those states. To summarize the contribution of each regime variable, we also compute two
marginal contrasts from the four-cell decomposition. The marginal reservoir
effect is defined as the average sMAPE in low-reservoir cells minus the average
sMAPE in high-reservoir cells, so that positive values indicate that
low-reservoir conditions are harder to predict. The marginal gas effect is
defined as the average sMAPE in high-TTF cells minus the average sMAPE in
low-TTF cells, so that positive values indicate reduced forecast accuracy when
gas prices are high. This framework provides an interpretable way to evaluate whether model errors
are concentrated in specific hydro-gas market states, and whether such regime
dependence is already captured by lagged prices and calendar structure or
requires the broader feature set used by the full LightGBM model.

\section{Results}
\label{sec:results}

\subsection{Benchmark performance across all zones}
\label{subsec:benchmark}

\begin{figure}[!htb]
  \centering
  \includegraphics[width=\linewidth]{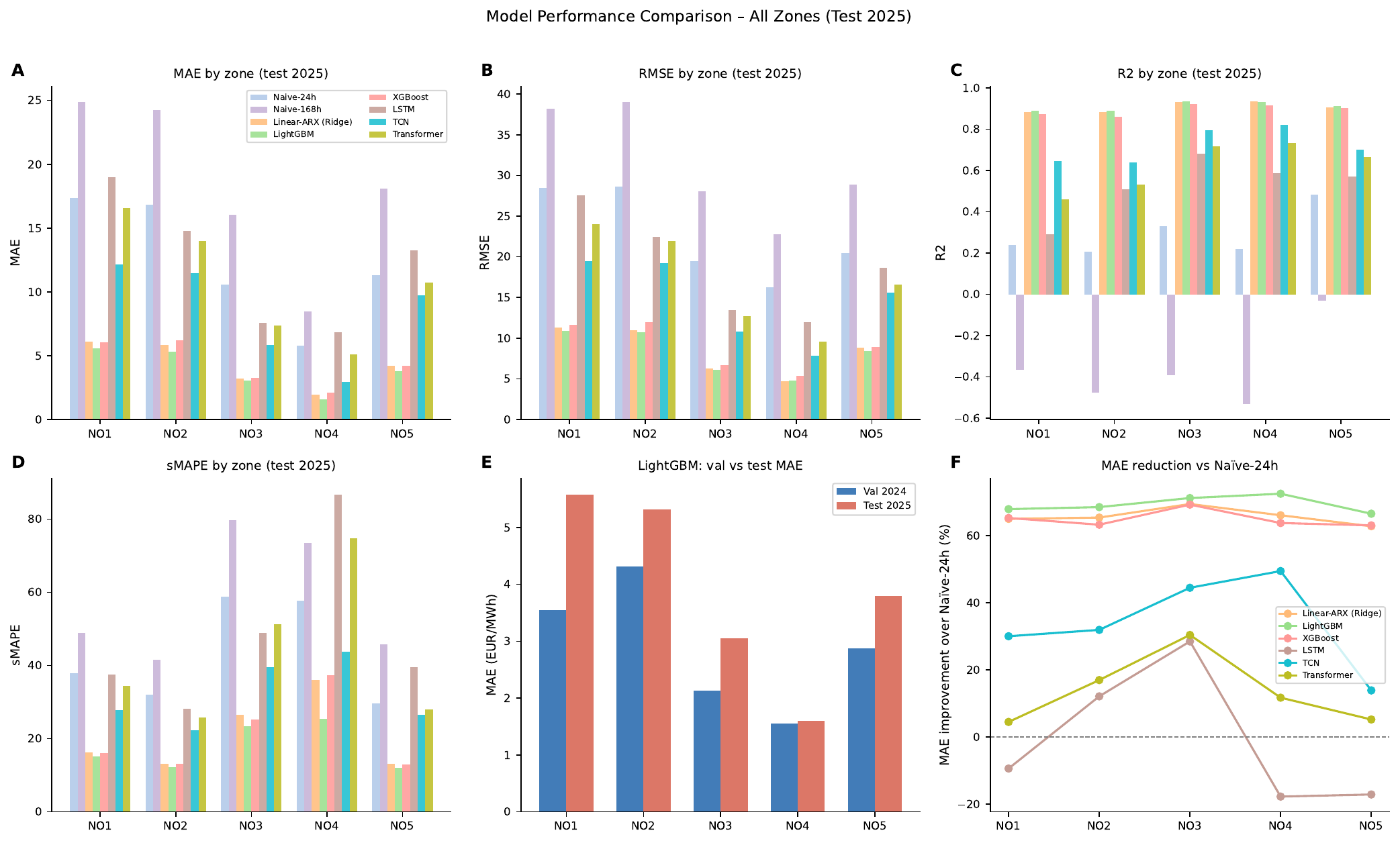}
  \caption{Model comparison across all zones (test set, 2025).
    \textbf{(A)}~MAE by zone (grouped bars, all eight models);
    \textbf{(B)}~RMSE by zone (grouped bars);
    \textbf{(C)}~$R^2$ by zone (grouped bars);
    \textbf{(D)}~sMAPE by zone (grouped bars);
    \textbf{(E)}~LightGBM validation (2024) vs.\ test (2025) MAE
    per zone, showing strong generalisation;
    \textbf{(F)}~MAE improvement over Na\"{i}ve-24h (\%) per
    zone for all non-na\"{i}ve models.
    LightGBM achieves the lowest MAE in all five zones
    (1.60-5.58~EUR/MWh). TCN is the best deep learning model
    (cross-zone mean MAE = 8.43 EUR/MWh). The performance gap
    between machine learning and deep learning is larger in NO1/NO2/NO5 than in
    the hydro-dominated NO3/NO4.}
  \label{fig:multizone_bar}
\end{figure}

Supplementary Table~S4 reports test-set (2025) results for all
models and zones; cross-zone summary means for all models are
given in Table~S5. LightGBM achieves the lowest MAE in every zone (ranging from 1.60~EUR/MWh in NO4 to 5.58~EUR/MWh in NO1, with a
cross-zone mean of 3.87~EUR/MWh and mean $R^2 = 0.911$), Ridge ARX is a
close second, while
XGBoost trails LightGBM in all zones. 
LightGBM reduces MAE by 66-72\% over the Na\"{i}ve-24h baseline, while XGBoost reduces it by 63-69\%. The Na\"{i}ve-168h model
underperforms the Na\"{i}ve-24h in every zone, confirming that within-week
price patterns shift too rapidly after the energy crisis to be captured
by a fixed weekly repetition. Among the three deep learning architectures,
the TCN achieves the strongest performance and outperforms Na\"{i}ve-24h in all five zones.
The Transformer also beats Na\"{i}ve-24h in all five zones, whereas the LSTM does so only in NO2 and NO3.
All three deep learning models remain substantially behind LightGBM
in the price-volatile southern zones. Figure~\ref{fig:multizone_bar} provides the full multi-metric comparison across all zones and models.

The lower absolute MAE in NO3 and NO4 reflects their structurally
lower price levels. The 2025 means are 21.0 and 8.6 EUR/MWh respectively
rather than higher proportional accuracy. The sMAPE values confirm
that NO3 and NO4 are not easier to forecast on a relative basis
(sMAPE 23-36\% vs 12-16\% for NO1/NO2/NO5). In NO4, Ridge ARX 
achieves a marginally higher $R^2$ (0.934) than LightGBM (0.933) 
despite a higher MAE (1.969 vs.\ 1.598~EUR/MWh); this shows the 
disproportionate influence of a small number of extreme-spike hours 
on the $R^2$ denominator in a low-price zone, and does not alter the 
MAE-based ranking. Per-zone MAE
and $R^2$ scatter, MAE reduction heatmap, and validation-to-test
transfer plots are provided in Supplementary Figure~S3; a full
heatmap of all four metrics across all models and zones is in
Supplementary Figure~S4.

\subsection{Statistical significance in Diebold-Mariano tests}
\label{subsec:dm}

\begin{figure}[!htb]
  \centering
  \includegraphics[width=\linewidth]{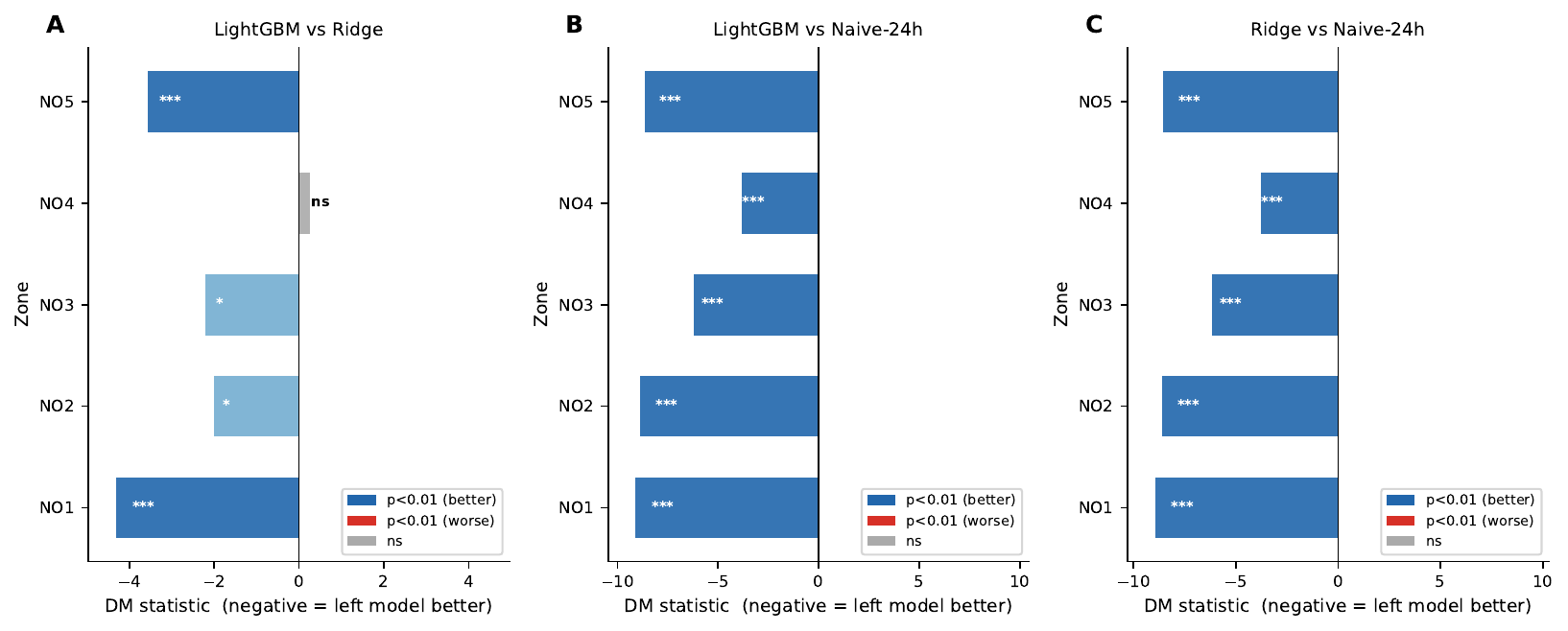}
  \caption{Diebold-Mariano pairwise test results (HLN-corrected,
    one-sided, 23-lag HAC variance correction), all five zones.
    \textbf{(A)}~LightGBM vs.~Ridge ARX by zone;
    \textbf{(B)}~LightGBM vs.~Na\"{i}ve-24h by zone;
    \textbf{(C)}~Ridge ARX vs.~Na\"{i}ve-24h by zone.
    Bars show the DM statistic and embedded markers denote
    significance. Negative values imply superior accuracy of the left model.
    Both machine learning models significantly beat Na\"{i}ve-24h ($p{<}0.001$)
    in all zones. LightGBM significantly outperforms Ridge in
    NO1 ($^{***}$), NO2 ($^{*}$), NO3 ($^{*}$), and NO5 ($^{***}$);
    not significant in NO4.}
  \label{fig:dm_heatmap}
\end{figure}

Figure~\ref{fig:dm_heatmap} and
Table~S6 summarise 
pairwise Diebold-Mariano tests (HLN-corrected, one-sided, with a 23-lag HAC variance correction). 
Both LightGBM and Ridge ARX are statistically significantly better 
than Na\"{i}ve-24h across all five zones. LightGBM significantly 
outperforms Ridge ARX in NO1, NO2, 
NO3, and NO5, but the 
advantage is not statistically significant in 
NO4.

The contrast is clearest in NO4, a northern, 
hydro-dominated zone with lower price volatility and simpler price 
dynamics. Autocorrelation at lag $168\,h$ is $0.65$ and $0.64$ for 
NO3 and NO4, compared with $0.81$-$0.82$ for NO1, NO2, and NO5. In 
NO4, a well-engineered linear model captures most of the available 
predictable structure, leaving no statistically detectable gap to 
LightGBM. NO3 is also hydro-dominated, but it still 
shows a smaller yet statistically significant LightGBM advantage. In 
the more volatile southern zones, where price formation is more 
tightly coupled to continental gas markets and intraday demand 
spikes, the non-linear flexibility of LightGBM delivers a 
clear improvement.

The static DM test, conducted on the full test set, is 
complemented by the rolling-origin backtest in 
section~\ref{subsec:walkforward}, which shows that LightGBM's 
superiority over Na\"{i}ve-24h holds in every single week of 
2025 across all zones (260/260 steps).

\subsection{Walk-Forward rolling-origin backtest}
\label{subsec:walkforward}

\begin{figure}[!htb]
  \centering
  \includegraphics[width=\linewidth]{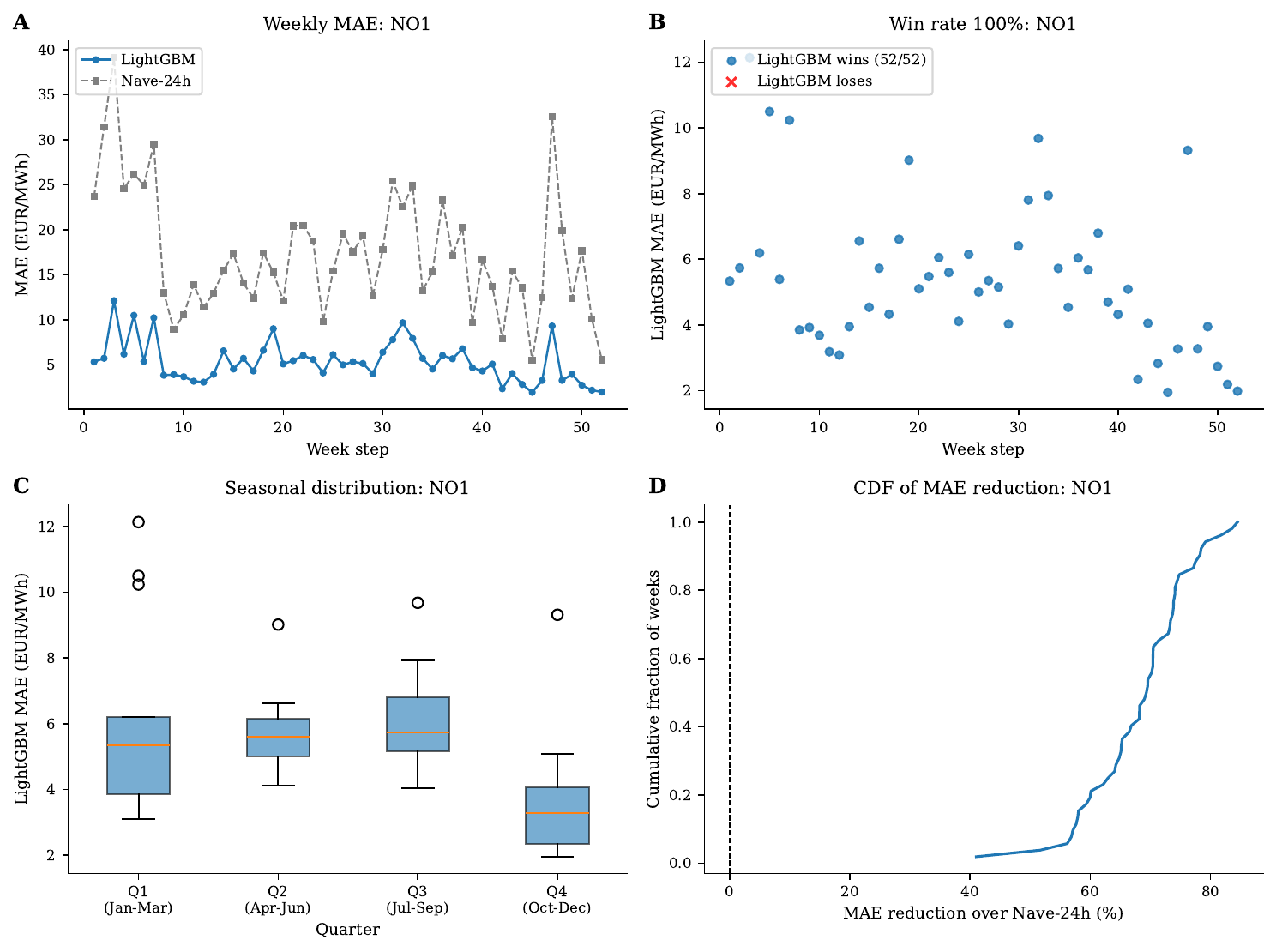}
  \caption{Walk-forward rolling-origin backtest for NO1
    (52 weekly steps, 2025).
    \textbf{(A)}~Weekly MAE for LightGBM and Na\"{i}ve-24h baselines;
    \textbf{(B)}~Weekly win/loss scatter for LightGBM relative to
    Na\"{i}ve-24h (52/52 wins; NO1 shown as representative zone);
    \textbf{(C)}~Seasonal distribution of weekly LightGBM MAE
    by quarter;
    \textbf{(D)}~Cumulative distribution of weekly MAE reduction
    over Na\"{i}ve-24h.
    Winter weeks with cold-snap spikes show the highest errors.
    Cross-zone summary statistics are in Supplementary Figure~S2.}
  \label{fig:walkforward}
\end{figure}

Figure~\ref{fig:walkforward} shows the week by week walk forward backtest for NO1 across all 52 evaluation steps, including seasonal breakdowns and cumulative win rate distributions; Supplementary Figure~S2 provides the cross zone summary. LightGBM outperforms Naïve-24h in every week across all five zones (260/260 steps), reducing mean weekly MAE by 68 to 73\% relative to the naive baseline. Per zone mean weekly MAE is 5.36, 5.03, 2.99, 1.57, and 3.62~EUR/MWh for NO1 to NO5, respectively, consistent with the static test set results in Supplementary Table~S4.

For NO1, the highest forecast errors occur in winter cold-snap weeks, especially in mid-January, with a secondary cluster in late November. Autumn weeks outside those spikes exhibit the lowest errors. This seasonal pattern is consistent with the regime analysis in Section~\ref{subsec:regime}, as winter and late autumn weeks coincide with high TTF and tighter hydro conditions under which forecast errors are elevated. The rolling origin design demonstrates that the superior performance of LightGBM is not driven by a single favourable test window, but holds consistently across all market conditions observed in 2025.

\subsection{Pre-crisis versus post-crisis data}

Table~\ref{tab:PrePostCrisis} compares LightGBM across three training windows. The full 2019-2023 model gives the best MAE and sMAPE in every zone. It also gives the best RMSE and $R^2$ in four of five zones; the exception is NO4, where the post-crisis-only window marginally outperforms on those two metrics ($\Delta\mathrm{RMSE}=0.09$, $\Delta R^2=0.002$). This shows that pre-crisis and post-crisis observations are most useful when they are used together. Training only on 2019-2021 is weaker in most zone-metric combinations. Pre-crisis data alone therefore do not describe the current regime well, but they still add useful information when combined with post-crisis data.

The comparison between the two reduced windows is also informative. In NO2 and NO3, the post-crisis-only model outperforms the pre-crisis-only model across all four metrics. In NO1 and NO4, the post-crisis-only model improves MAE, RMSE, and $R^2$, but not sMAPE. In NO5 the pattern is mixed. The pre-crisis-only model gives lower MAE and sMAPE, whereas the post-crisis-only model gives lower RMSE and higher $R^2$. Overall, the results show that post-crisis data are necessary to capture the current market regime, while the full 2019-2023 window remains best because it also preserves broader seasonal and autoregressive structure.

\begin{table}[!htbp]
\caption{LightGBM performance under three training windows on the 2025 test set after model selection on 2024.}
\label{tab:PrePostCrisis}
\centering
{\scriptsize
\setlength{\tabcolsep}{3pt}
\renewcommand{\arraystretch}{1.15}
\resizebox{\textwidth}{!}{%
\begin{tabular}{lcccccccccccc}
\toprule
& \multicolumn{4}{c}{Train 2019-2021} & \multicolumn{4}{c}{Train 2022-2023} & \multicolumn{4}{c}{Train 2019-2023} \\
\cmidrule(lr){2-5} \cmidrule(lr){6-9} \cmidrule(lr){10-13}
Zone & MAE & RMSE & sMAPE & $R^2$ & MAE & RMSE & sMAPE & $R^2$ & MAE & RMSE & sMAPE & $R^2$ \\
\midrule
NO1 & 6.14 & 11.79 & 15.91 & 0.8699 & 6.10 & 11.30 & 16.22 & 0.8803 & \textbf{5.58} & \textbf{10.88} & \textbf{15.09} & \textbf{0.8890} \\
NO2 & 6.07 & 11.80 & 13.15 & 0.8649 & 5.43 & 10.76 & 11.80 & 0.8876 & \textbf{5.32} & \textbf{10.72} & \textbf{12.25} & \textbf{0.8885} \\
NO3 & 3.22 & 6.49 & 24.84 & 0.9257 & 3.13 & 6.22 & 24.31 & 0.9317 & \textbf{3.05} & \textbf{6.14} & \textbf{23.38} & \textbf{0.9335} \\
NO4 & 1.73 & 5.08 & 26.52 & 0.9238 & 1.66 & \textbf{4.68} & 27.43 & \textbf{0.9353} & \textbf{1.60} & 4.77 & \textbf{25.32} & 0.9328 \\
NO5 & 4.14 & 8.98 & 12.68 & 0.9003 & 4.23 & 8.57 & 13.20 & 0.9092 & \textbf{3.80} & \textbf{8.44} & \textbf{12.07} & \textbf{0.9120} \\
\bottomrule
\end{tabular}%
}
}
\end{table}

\subsection{Feature group ablation}
\label{subsec:ablation}

\begin{figure*}[!htb]
    \centering
    \includegraphics[width=\textwidth]{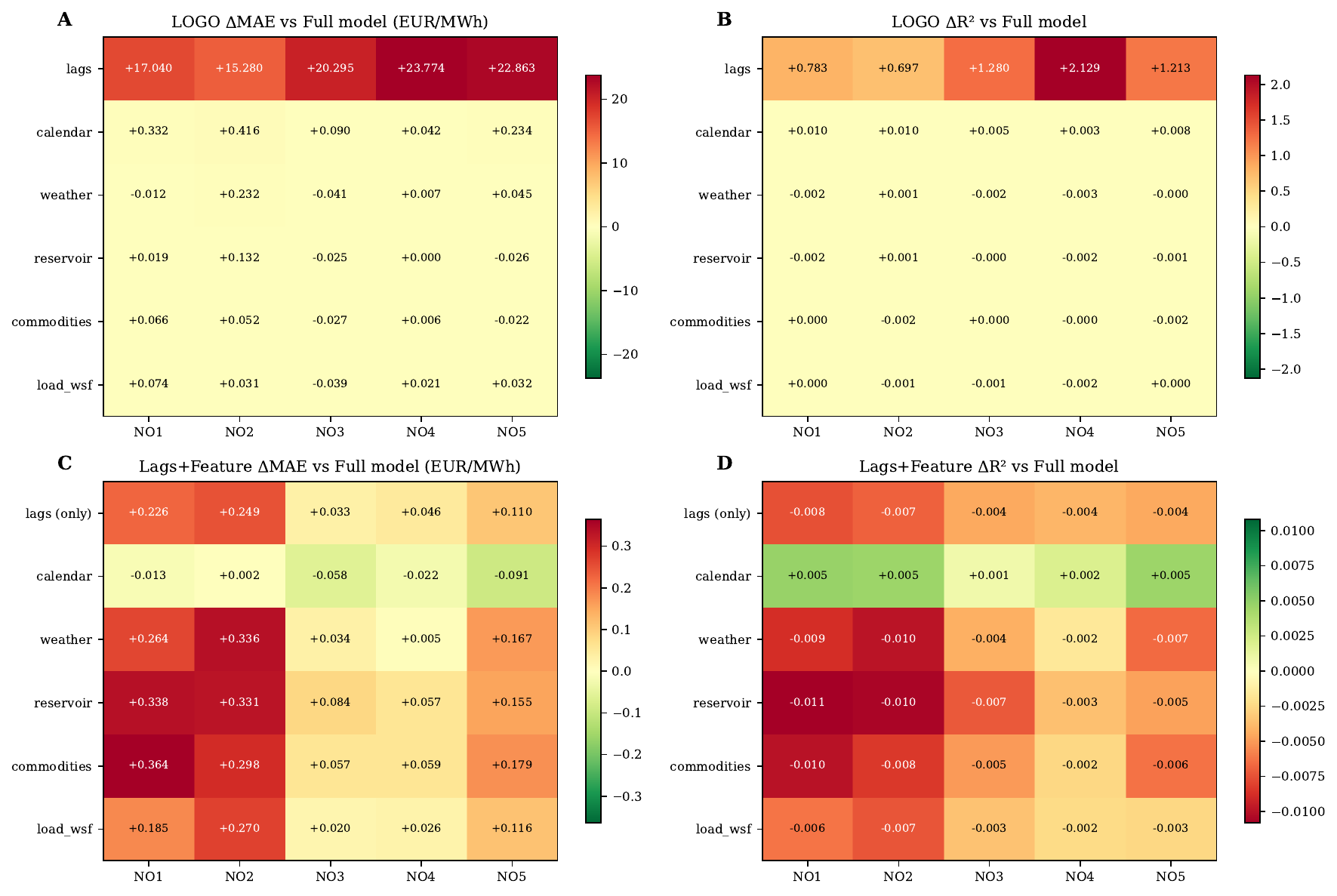}
    \caption{
    Feature group ablation across NO1-NO5.
    \textbf{(A)} LOGO $\Delta\mathrm{MAE}$ relative to the full model.
    \textbf{(B)} LOGO $\Delta R^2$ relative to the full model.
    \textbf{(C)} Lags-plus-one-group $\Delta\mathrm{MAE}$ relative to the full model.
    \textbf{(D)} Lags-plus-one-group $\Delta R^2$ relative to the full model}
    \label{fig:ablation}
\end{figure*}

Figure~\ref{fig:ablation} and Table~S7,S8 summarize the feature group ablation results across
all five bidding zones. In the leave-one-group-out analysis, removing the lag
group causes by far the largest deterioration in forecast accuracy in every
zone, with $\Delta\mathrm{MAE}=17.04$, $15.28$, $20.30$, $23.77$, and
$22.86$~EUR/MWh in NO1-NO5, respectively, and corresponding drops in
$R^2$ of $0.783$, $0.697$, $1.280$, $2.129$, and $1.213$. This consistently
shows that autoregressive price information is the dominant source of
predictive power in the model. Among the exogenous groups, calendar features provide the clearest and most
consistent marginal contribution. Removing the calendar group increases MAE by
$0.33$~EUR/MWh in NO1, $0.42$ in NO2, $0.09$ in NO3, $0.04$ in NO4, and
$0.23$ in NO5, with small and generally positive changes in $R^2$.
The effect is therefore strongest in NO1, NO2, and NO5, and much weaker in NO3
and NO4, indicating that calendar structure contributes useful information, but
its importance varies across bidding zones.

By contrast, the remaining exogenous groups have only weak marginal effects in
the full model. In the LOGO analysis, weather, reservoir, commodities, and
\texttt{load\_wsf} typically change MAE by only a few hundredths of EUR/MWh,
and the direction is not consistent across zones. This suggests that these
variables provide little additional information once lagged prices and calendar
effects are already included. A likely explanation is that much of the
short-horizon information contained in weather, hydrological conditions, fuel
markets, and system forecasts is already reflected in recent prices, so their
incremental contribution becomes small after the autoregressive structure has
been modeled.

The group-only experiments in Panels~C-D provide a complementary view by
measuring how well each feature group performs when added to lags alone.
Calendar remains the only exogenous group that consistently comes close to the
full model in every zone: the lags+calendar specification is within
$0.013$, $0.002$, $0.058$, $0.022$, and $0.091$~EUR/MWh of the full model in
NO1-NO5, respectively, and it slightly outperforms the full model in four of
the five zones. Outside calendar, no lags-plus-one-group specification matches
this consistency; most remain clearly worse than the full multimodal model.

Taken together, these results show a clear hierarchy of feature utility.
Lagged prices dominate forecasting performance in all zones, calendar variables
provide the most robust exogenous contribution, and the remaining
exogenous features add little unique predictive value once the core
autoregressive and seasonal structure is already represented in the model.

\subsection{Conditional regime analysis}
\label{subsec:regime}

\begin{figure}[!htb]
  \centering
  \includegraphics[width=\linewidth]{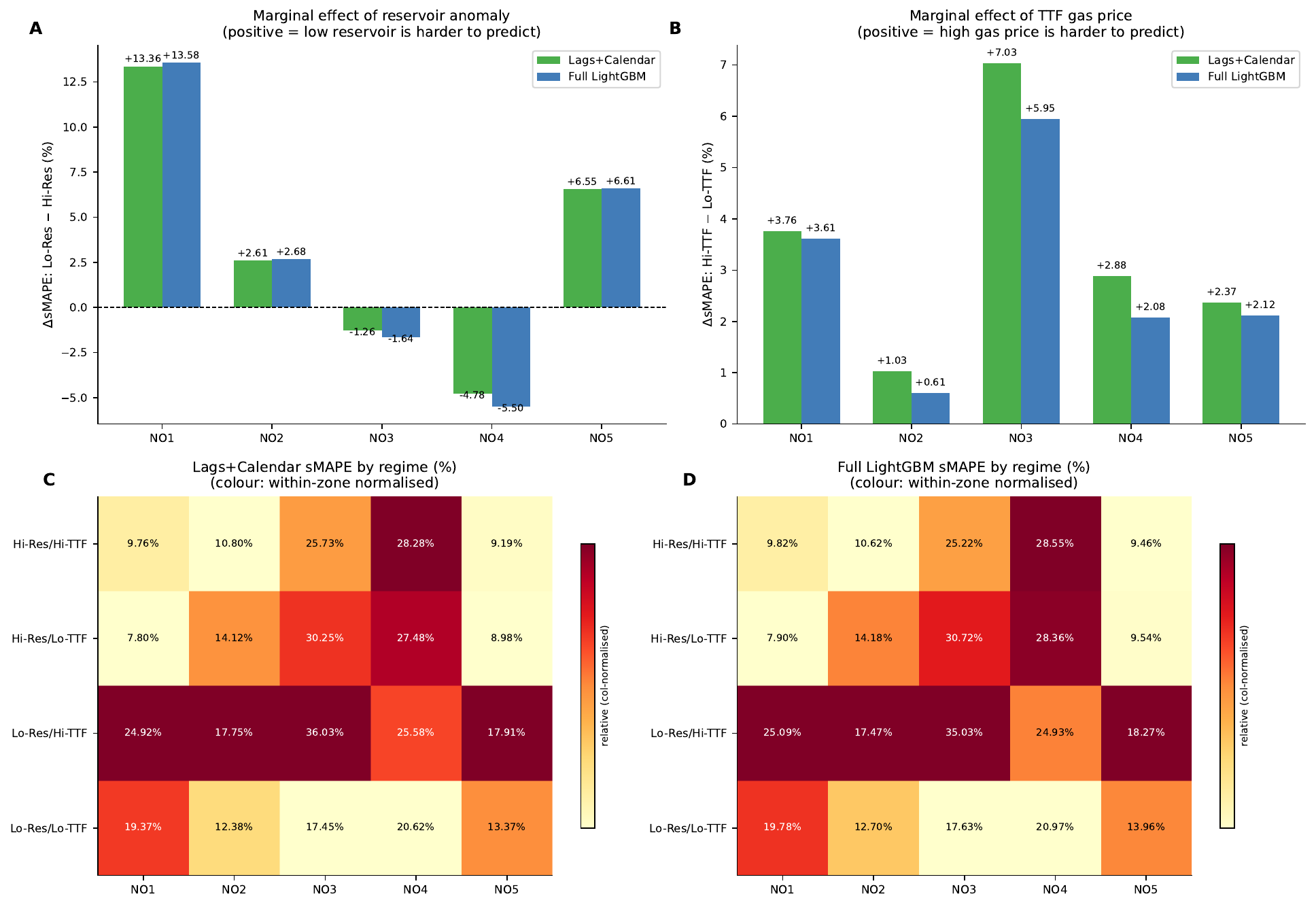}
  \caption{Conditional regime analysis across the five Norwegian bidding zones.
    \textbf{(A)}~Marginal effect of reservoir anomaly on sMAPE, computed as the
    average sMAPE in low-reservoir cells minus the average sMAPE in
    high-reservoir cells; positive values indicate that low-reservoir states
    are harder to predict.
    \textbf{(B)}~Marginal effect of TTF gas price on sMAPE, computed as the
    average sMAPE in high-TTF cells minus the average sMAPE in low-TTF cells;
    positive values indicate that high gas-price states are harder to predict.
    \textbf{(C)}~Lags+Calendar sMAPE in the four joint regime cells defined by
    reservoir anomaly (high/low) and TTF gas price (high/low).
    \textbf{(D)}~Full LightGBM sMAPE in the same regime cells.
    Cell labels report absolute sMAPE values, while colour intensity is
    normalised within each zone to highlight the relative difficulty of each
    regime.}
  \label{fig:regime}
\end{figure}

Figure~\ref{fig:regime} and Table~S9 summarize regime-dependent forecast accuracy across
all five bidding zones for the reduced Lags+Calendar model and the full
LightGBM model. Panel~A shows that the marginal effect of reservoir anomaly is
positive in NO1, NO2, and NO5, but negative in NO3 and NO4. For the full
LightGBM model, the low-minus-high reservoir contrast equals $+13.58$,
$+2.68$, $-1.64$, $-5.50$, and $+6.61$ percentage points in NO1-NO5,
respectively. The reduced Lags+Calendar model follows the same sign pattern.
This indicates that low-reservoir conditions are harder to predict in the
southern zones, whereas forecast errors are slightly higher under
high-reservoir conditions in NO3 and NO4.

Panel~B shows that high TTF prices consistently increase sMAPE in every zone.
For the full LightGBM model, the high-minus-low TTF contrast is $+3.61$,
$+0.61$, $+5.95$, $+2.08$, and $+2.12$ percentage points in NO1-NO5,
respectively, again with the reduced model showing the same direction. Thus,
high gas prices represent the most consistent source of forecast difficulty
across all bidding zones.

The regime heatmaps in Panels~C-D show that the low-reservoir/high-TTF regime
is among the most difficult states in NO1, NO2, NO3, and NO5. For the full
LightGBM model, sMAPE in this regime reaches $25.09\%$ in NO1, $17.47\%$ in
NO2, $35.03\%$ in NO3, and $18.27\%$ in NO5. In NO4, by contrast, the highest
error occurs under high-reservoir/high-TTF conditions, where sMAPE reaches
$28.55\%$ for Full LightGBM and $28.28\%$ for Lags+Calendar.

Across regime cells, the reduced Lags+Calendar model closely matches the full
LightGBM model and often performs slightly better. In NO1 it lowers sMAPE in
all four regime cells, for example from $9.82\%$ to $9.76\%$ in the
high-reservoir/high-TTF regime and from $25.09\%$ to $24.92\%$ in the
low-reservoir/high-TTF regime. The results demonstrate much of the information
needed for point forecasting is already captured by lagged prices and calendar
structure, limiting the incremental accuracy gains from the broader feature
set. At the same time, reservoir anomaly and TTF price remain important for
interpretation, as they clearly identify the hydro-thermal market states in
which forecast errors increase.

\subsection{Failure mode analysis of worst predictions}
\label{subsec:failures}

The 20 largest absolute errors on the NO1 test set are 
examined in Supplementary Figure~S5. Eighteen of the 20 
errors are under-predictions, meaning the model 
consistently forecast prices far below what occurred 
rather than over-shooting. The single largest error falls 
on 2025-01-20 at 08:00 on a Monday morning. The actual 
price reached 448~EUR/MWh while LightGBM predicted 
229~EUR/MWh, yielding an absolute error of 219.30~EUR/MWh. At this timestamp the reservoir anomaly was 
near zero, TTF was 46.9~EUR/MWh, and the 24-hour price 
lag was only 40~EUR/MWh, so the model had no feature 
signal indicating a spike of this magnitude was imminent.

This event is representative rather than exceptional. 
Sixteen of the 20 errors fall on Monday or Tuesday, when 
prices can jump sharply from a quiet weekend level with 
no prior-day lag to warn the model. Nine of the 20 
errors occur in the morning peak hours (06:00-09:00), 
when demand rises steeply and the lag-based signal is 
at its weakest relative to the actual price movement. 
In every case, feature diagnostics confirm the same 
combination of elevated TTF and near-zero reservoir 
anomaly, matching precisely the high-error regime cells 
identified in Section~\ref{subsec:regime}. Taken 
together, these patterns show that the model's worst 
failures are not random but are concentrated in a 
specific and identifiable market condition. Specifically, the model struggles when a gas-driven price spike occurs on a Monday or Tuesday morning after a calm weekend, with no reservoir scarcity signal and no recent lag precedent.

\section{Discussion}

\subsection{Pre-crisis versus post-crisis data}

The training-window comparison shows that post-crisis data are necessary but not sufficient. Models trained only on 2022-2023 generally outperform models trained only on 2019-2021, which confirms that pre-crisis data alone do not represent the current regime well. However, the full 2019-2023 window gives the best performance in most zone-metric combinations in Table~\ref{tab:PrePostCrisis}, with the only exception being NO4 where the post-crisis-only window marginally outperforms on RMSE and $R^2$. This indicates that pre-crisis observations still contain useful information when they are combined with post-crisis data.

A likely explanation is that the pre-crisis sample helps the model learn stable seasonal and autoregressive structure, while the post-crisis sample is needed to represent the new hydro-gas price regime. The main implication is therefore not that older data should be discarded, but that they should be used together with enough recent data from the current market state.

\subsection{Autoregression and exogenous information}

The feature ablation results show that autoregressive structure remains the
dominant source of predictive power in Norwegian next-hour forecasting of the
Nord Pool day-ahead electricity price. Removing lag features increases MAE by 15.28-23.77~EUR/MWh across
NO1-NO5, whereas removing any exogenous group changes MAE by only a few
hundredths of EUR/MWh in most cases. Calendar variables are the only exogenous
group that provides a clear and consistent gain across all zones, but even this
effect remains modest relative to the contribution of lagged prices. This pattern does not imply that reservoir, weather, commodity, and system
forecast variables are uninformative. Rather, it suggests that much of their
short-horizon predictive content is already encoded in recent prices, which act
as a compressed summary of market responses to these fundamentals. In that
sense, internal correlation with lagged prices limits the incremental value of
many exogenous predictors for point forecasting accuracy. At the same time, the regime analysis shows that multimodal variables remain
important for diagnosis and interpretation. Reservoir anomaly and TTF gas price
do not materially improve average point accuracy beyond the lag and calendar
baseline, but they clearly stratify the hydro-thermal states in which forecast
errors increase. Exogenous variables therefore retain substantial explanatory
value even when their marginal contribution to test-set MAE is small.

\subsection{Zone heterogeneity}

The benchmark and significance results reveal a clear difference between the
southern zones (NO1, NO2, NO5) and the northern zones (NO3, NO4). LightGBM
achieves the lowest MAE in every zone, but its advantage over Ridge ARX is
statistically significant in NO1, NO2, NO3, and NO5. NO4 is the only zone in
which the two models perform similarly enough that the difference is not
statistically significant, indicating that a well-specified linear model already
captures most of the predictable structure in that market. The ablation and regime results reinforce this heterogeneity. Calendar effects
are strongest in NO1, NO2, and NO5, while the marginal effect of reservoir
state differs by zone. Low-reservoir conditions are harder to predict in NO1,
NO2, and NO5, but forecast errors are slightly higher under high-reservoir
conditions in NO3 and NO4. By contrast, high TTF prices increase forecast
difficulty in all five zones, making gas stress the most consistent adverse
regime across the system. Taken together, these results indicate that model complexity is most useful in
zones where market conditions are more strongly associated with non-linear and
externally coupled dynamics. In zones where price formation is more regular,
the practical advantage of LightGBM over Ridge ARX becomes much smaller even
though LightGBM still ranks first by MAE.

\subsection{Robustness across time and regimes}

The rolling-origin backtest shows that LightGBM outperforms Na\"{i}ve-24h in
every week of 2025 across all five bidding zones, confirming that its
superiority is not driven by a narrow subset of the test period. This temporal
stability strengthens the benchmark findings and supports the use of the model
under a wide range of post-crisis market conditions. The regime analysis adds an important qualification to this result. Although
the reduced Lags+Calendar model closely matches the full LightGBM model across
regime cells and often performs slightly better, forecast errors still increase
markedly in specific hydro-gas states, especially under high TTF conditions and
in low-reservoir states in the southern zones. This indicates that the main
forecast signal is already captured by autoregressive and seasonal structure,
but market-state variables remain essential for understanding when prediction
becomes more difficult. The failure analysis is consistent with this interpretation. The largest
forecast errors cluster around sharp upward price movements, often on Monday or
Tuesday mornings, when recent lags provide little warning of an abrupt regime
shift. These events show the practical boundary of lag dominated forecasting.
Models can perform strongly on average while still remaining vulnerable to rare
and fast-moving price spikes.

\subsection{Limitations}

Several limitations should be noted. First, the deep learning models are
trained with basic optimization; more delicate architectural optimisation may narrow the gap to LightGBM. Second, the study focuses
on point forecasting, whereas probabilistic forecasts would be more informative
for operational decision-making under extreme price uncertainty.

Third, weather inputs are represented at zone level using a limited spatial
summary, which may not fully capture local demand- or supply-relevant
conditions. Fourth, although the feature ablation and regime analyses are
extended to all five zones, they remain descriptive diagnostics and do not by
themselves establish causal relationships between hydro, gas, and forecast
error. Finally, the present analysis evaluates a fixed set of models and
feature groups; alternative formulations, including regime-specific models or
more explicit interactions between hydro and gas variables, may provide further
gains.

\section{Conclusion}

This study provides a multi-zone benchmark for hourly one-step-ahead forecasting
of the Nord Pool day-ahead electricity price across all five Norwegian bidding
zones on a common 2025 test period. LightGBM achieves the lowest MAE in every
zone, with errors ranging from 1.60~EUR/MWh in NO4 to 5.58~EUR/MWh in NO1 and a
cross-zone mean $R^2 = 0.911$. Its advantage over Na\"{i}ve-24h is large and
consistent in both static and rolling-origin evaluation, and it is
statistically significant against Ridge ARX in NO1, NO2, NO3, and NO5, with
NO4 the only non-significant zone.

While the day-ahead market is the primary trading horizon in Nord Pool, the intraday market remains an active and operationally important layer of system operation, enabling continuous trading on hourly and 15-minute products \cite{NordPool_intraday}. This operational structure makes short-term forecasting relevant for near-delivery adjustments, operational planning, and risk management.

The main methodological finding is that lagged prices dominate predictive
performance. Feature ablation shows that removing lag features causes a large
loss of accuracy in every zone, whereas most exogenous groups provide only
small marginal gains once lagged prices are already included. Calendar
variables are the most robust secondary feature group, while reservoir,
commodity, weather, and load or renewable forecast variables contribute little
additional improvement to average point accuracy. At the same time, exogenous variables remain important for understanding model behaviour. The conditional regime analysis shows that forecast errors increase
most consistently under high TTF gas-price conditions, while the effect of
reservoir state differs across bidding zones. This indicates that multimodal
features may add limited incremental value for prediction, yet still play an
important role in diagnosis by identifying the hydro-thermal market states in
which models struggle most.

Overall, the results suggest a practical division of labour between features.
Lagged prices and calendar effects carry most of the information needed for
accurate next-hour point forecasts of the day-ahead price, whereas broader
exogenous variables are most valuable for interpreting forecast risk and model
failure modes. Future
work may therefore focus not only on improving average accuracy, but also on
developing regime-aware or probabilistic models that better characterize the
conditions under which forecast errors become large.

\section*{Data Availability}
All data collection scripts, preprocessing pipelines, model code,
and numerical results are available at \url{https://github.com/myptd/NOR_EPF}.
Primary data sources include ENTSO-E Transparency Platform (attribution
required per Terms of Use), Open-Meteo Historical Weather Archive
application programming interface (API) \citep{openmeteo}, NVE Magasinstatistikk application programming interface under the Norwegian Licence for Open Government Data (NLOD),
Yahoo Finance via \texttt{yfinance}.

\section*{Contributor Roles Taxonomy (CRediT) Author Statement}
MTDP: Conceptualisation, Data Curation, Methodology,
Software, Formal Analysis, Visualisation, Writing original draft.

TTT: Data Curation, Validation, Writing review and editing.

HPH: Data Curation, Validation, Writing review and editing.

DTN: Supervision, Methodology, Software, Validation,
Writing review and editing.

\section*{Use of AI software}

Large language models were used to improve the wording and grammar of some texts but not to generate new content.

\section*{Declaration of Competing Interests}
The authors declare no competing interests.

\section*{Acknowledgements}

The authors acknowledge the use of Sigma2 computational resources through the University of Oslo project \textbf{nn9780k}. The authors also thank Nico Hacker for his correspondence regarding the clarification of prediction horizon.

\section*{Funding}

This work received no specific funding.

\bibliographystyle{elsarticle-num}
\bibliography{references}

\appendix


\section*{Supplementary material}

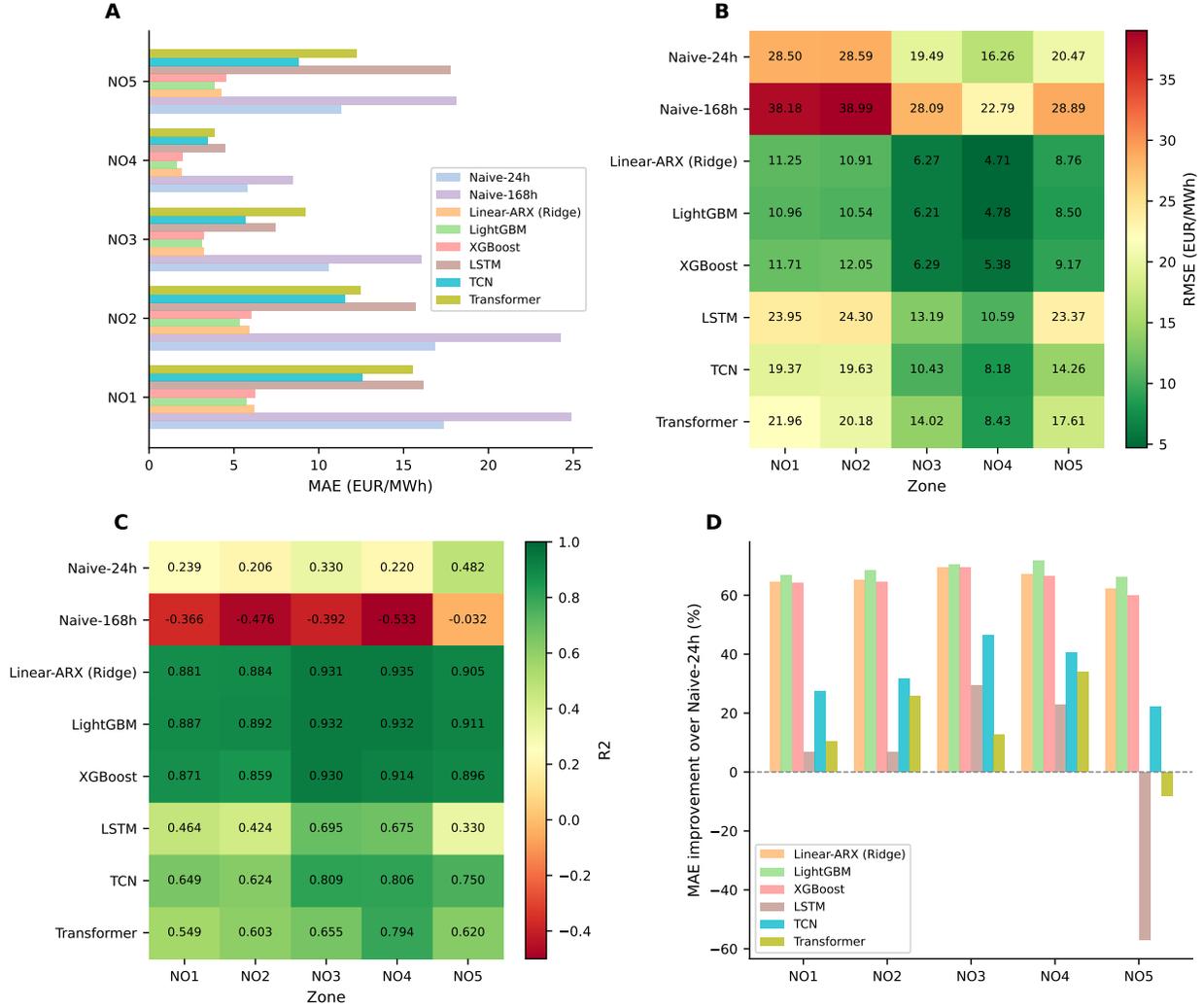

Figure S4: Multi-zone model comparison (test set, 2025). Colour-coded performance metrics for all five bidding zones (NO1–NO5) and eight forecasting models (Naïve-24h, Naïve-168h, Ridge ARX, LightGBM, XGBoost, LSTM, TCN, Transformer). **(A)** MAE (EUR/MWh); **(B)** RMSE (EUR/MWh); **(C)** $R^2$; **(D)** sMAPE (%). Gradient-boosted trees (LightGBM, XGBoost) consistently achieve the lowest MAE and highest $R^2$ across all zones; deep learning models (LSTM, TCN, Transformer) substantially underperform relative to gradient-boosted trees and Ridge ARX.

4

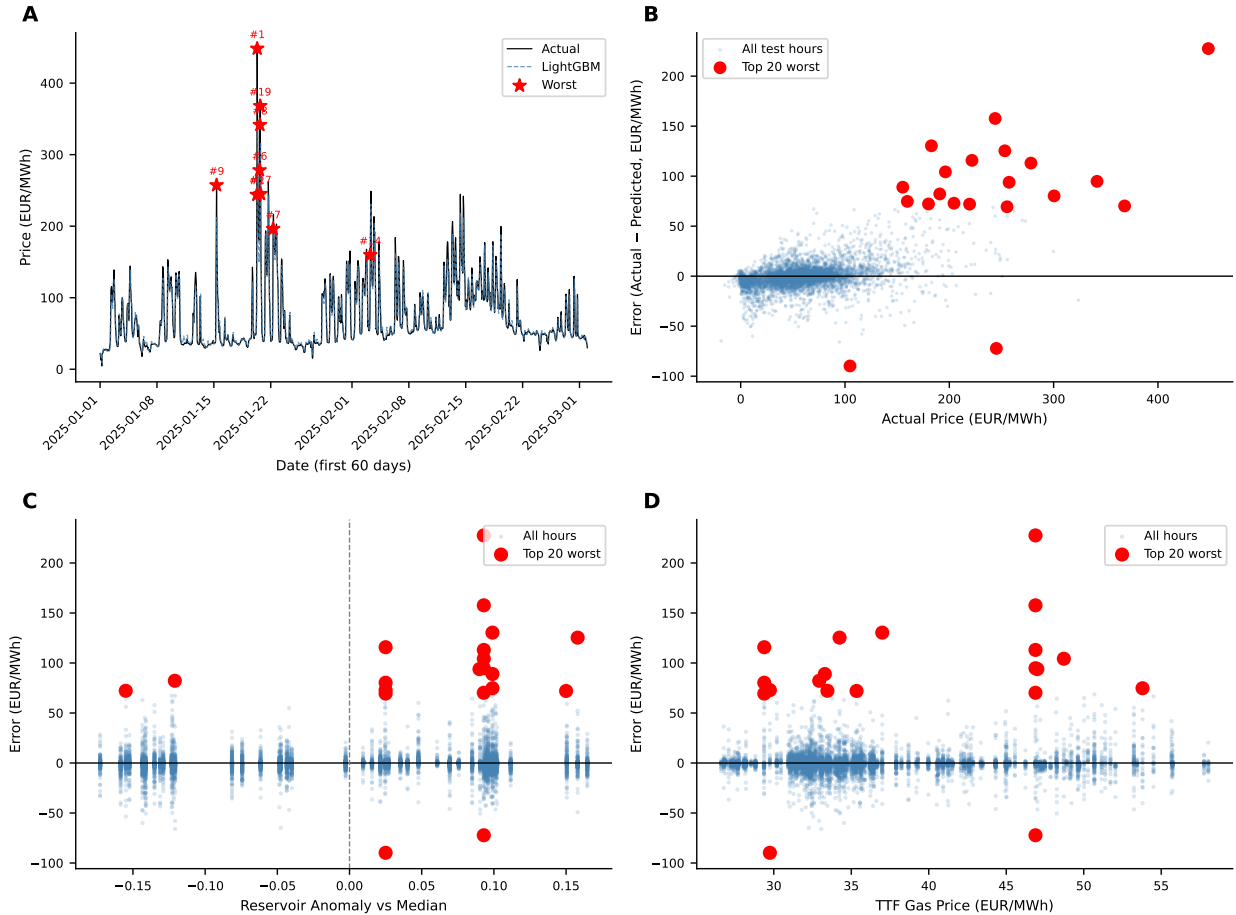

Figure S5: Feature diagnostics at worst-prediction timestamps (NO1 test set, 2025). **(A)** Time series of the first 60 test days with the 20 largest absolute errors marked by vertical lines; **(B)** Actual vs. predicted scatter for the 20 worst-error hours, annotated with rank; **(C)** Error magnitude by hour-of-day for the 20 worst predictions, showing concentration in morning peak hours (06:00–09:00); **(D)** Feature values (reservoir anomaly, TTF gas, load forecast) at the worst-error timestamps, illustrating that the January 20 cold-snap event (actual = 448 EUR/MWh, predicted = 221 EUR/MWh) coincided with near-zero reservoir anomaly and elevated TTF = 46.9 EUR/MWh, providing no ex-ante price-spike signal. Panels C and D are the primary diagnostic panels; panels A–B are included for temporal context.

# Supplementary Data

Table S1: Data sources used in this study.

| Source | Variables | Resolution | Period |
|---|---|---|---|
| ENTSO-E Transparency | Day-ahead prices, load forecast, wind/solar forecast, cross-border flows | Hourly | 2019–2025 |
| Open-Meteo Archive API | Temperature, wind speed, precipitation, radiation, soil moisture (20 variables) | Hourly | 2019–2025 |
| NVE Magasinstatistikk | Reservoir filling level, capacity, seasonal anomaly, trend features | Weekly | 1995–2025 |
| Yahoo Finance | TTF gas, European Union Allowance (EUA) carbon, Brent oil, EUR/NOK foreign exchange (FX), momentum indicators | Daily | 2010–2025 |

Table S2: Descriptive statistics of hourly day-ahead prices (EUR/MWh) by bidding zone, full study period 2019-01-01 to 2025-12-30 ($N = 61{,}338$ per zone). Skewness and excess kurtosis computed on the full hourly series. Neg.% denotes the fraction of hours with price $< 0$.

| Zone | $N$ | Mean | SD | Median | Min | Max | Skew | Kurt | Neg.% |
|---|---|---|---|---|---|---|---|---|---|
| NO1 | 61,338 | 69.0 | 74.9 | 47.3 | −61.8 | 799.9 | 2.91 | 12.4 | 1.11 |
| NO2 | 61,338 | 75.7 | 81.6 | 52.9 | −61.8 | 898.3 | 3.08 | 13.1 | 0.69 |
| NO3 | 61,338 | 31.2 | 35.0 | 24.5 | −14.9 | 590.0 | 4.97 | 43.8 | 1.10 |
| NO4 | 61,338 | 24.1 | 25.9 | 18.9 | −24.1 | 504.8 | 4.99 | 54.3 | 1.12 |
| NO5 | 61,338 | 67.1 | 74.4 | 44.5 | −21.7 | 800.0 | 2.97 | 12.8 | 0.90 |

Table S3: Feature groups used in modelling.

| Group | Key features | Availability |
|---|---|---|
| Lags | $p_{t-1}, p_{t-2}, p_{t-6}, p_{t-12}, p_{t-24}, p_{t-48}, p_{t-168}$; rolling 24h and 168h mean/std | Historical |
| Calendar | sin / cos encodings of hour, day-of-week, month | Known |
| Load/Wind-Solar (WSF) | Load forecast MW, wind/solar forecast | D-1 published |
| Weather | Temperature, wind speed, precipitation, radiation | NWP D-1 |
| Reservoir | Fill %, anomaly vs 20-yr median, trend features | Weekly |
| Commodities | TTF gas, EUA carbon, Brent, EUR/NOK, ratios | Settlement D-1 |

Table S4: Test-set results (2025), all zones and models. MAE and RMSE in EUR/MWh. Best ML result per zone in bold. Deep learning models (LSTM, TCN, Transformer) trained independently per zone (6 layers, hidden = 128, seq_len = 168 h, L1 loss).

| Zone | Model | MAE | RMSE | sMAPE | $R^2$ |
|---|---|---|---|---|---|
| NO1 | Naïve-24h | 17.34 | 28.50 | 37.9 | 0.239 |
| | Naïve-168h | 24.86 | 38.18 | 48.9 | −0.366 |
| | Ridge ARX | 6.17 | 11.25 | 16.4 | 0.881 |
| | **LightGBM** | **5.74** | **10.96** | **15.6** | **0.887** |
| | XGBoost | 6.24 | 11.71 | 16.7 | 0.871 |
| | LSTM | 16.18 | 23.95 | 34.6 | 0.464 |
| | TCN | 12.59 | 19.37 | 28.3 | 0.649 |
| | Transformer | 15.55 | 21.96 | 33.9 | 0.549 |
| NO2 | Naïve-24h | 16.83 | 28.59 | 32.0 | 0.207 |
| | Naïve-168h | 24.25 | 38.99 | 41.5 | −0.476 |
| | Ridge ARX | 5.89 | 10.91 | 13.3 | 0.884 |
| | **LightGBM** | **5.32** | **10.54** | **12.1** | **0.892** |
| | XGBoost | 6.00 | 12.05 | 13.0 | 0.859 |
| | LSTM | 15.70 | 24.30 | 28.0 | 0.424 |
| | TCN | 11.53 | 19.63 | 22.2 | 0.624 |
| | Transformer | 12.48 | 20.18 | 23.8 | 0.603 |
| NO3 | Naïve-24h | 10.55 | 19.49 | 58.9 | 0.330 |
| | Naïve-168h | 16.04 | 28.09 | 79.8 | −0.392 |
| | Ridge ARX | 3.25 | 6.27 | 26.6 | 0.931 |
| | **LightGBM** | **3.12** | **6.21** | **23.8** | **0.932** |

*(Table S4 continued)*

| Zone | Model | MAE | RMSE | sMAPE | $R^2$ |
|---|---|---|---|---|---|
| | XGBoost | 3.24 | 6.29 | 26.1 | 0.930 |
| | LSTM | 7.44 | 13.19 | 62.8 | 0.695 |
| | TCN | 5.67 | 10.43 | 38.9 | 0.809 |
| | Transformer | 9.22 | 14.02 | 93.7 | 0.655 |
| NO4 | Naïve-24h | 5.79 | 16.26 | 57.7 | 0.220 |
| | Naïve-168h | 8.49 | 22.79 | 73.4 | −0.533 |
| | Ridge ARX | 1.90 | 4.71 | 35.5 | 0.935 |
| | **LightGBM** | **1.64** | **4.78** | **25.4** | **0.933** |
| | XGBoost | 1.95 | 5.38 | 29.9 | 0.915 |
| | LSTM | 4.48 | 10.59 | 70.7 | 0.675 |
| | TCN | 3.44 | 8.18 | 82.5 | 0.806 |
| | Transformer | 3.82 | 8.43 | 76.6 | 0.794 |
| NO5 | Naïve-24h | 11.31 | 20.47 | 29.6 | 0.482 |
| | Naïve-168h | 18.11 | 28.89 | 45.8 | −0.032 |
| | Ridge ARX | 4.26 | 8.76 | 13.1 | 0.905 |
| | **LightGBM** | **3.84** | **8.50** | **12.0** | **0.911** |
| | XGBoost | 4.55 | 9.17 | 13.9 | 0.896 |
| | LSTM | 17.76 | 23.37 | 62.1 | 0.330 |
| | TCN | 8.81 | 14.26 | 24.0 | 0.751 |
| | Transformer | 12.22 | 17.61 | 35.7 | 0.620 |

Table S5: Cross-zone mean test-set performance (2025), ML models only. Naïve baselines included for reference. Arithmetic mean across NO1–NO5.

| Model | MAE | RMSE | sMAPE | $R^2$ |
|---|---|---|---|---|
| Naïve-24h | 12.37 | 22.66 | 43.2 | 0.296 |
| Naïve-168h | 18.35 | 31.39 | 57.9 | −0.360 |
| Ridge ARX | 4.29 | 8.38 | 21.0 | 0.907 |
| **LightGBM** | **3.93** | **8.20** | **17.8** | **0.911** |
| XGBoost | 4.40 | 8.92 | 19.9 | 0.894 |
| LSTM | 12.31 | 19.08 | 51.6 | 0.518 |
| TCN | 8.41 | 14.38 | 39.2 | 0.728 |
| Transformer | 10.66 | 16.44 | 52.7 | 0.644 |

Table S6: Pairwise Diebold–Mariano tests (HLN-corrected, one-sided, $h = 24$). Negative DM stat indicates the left model is more accurate. Significance levels: $^{*}p < 0.05$, $^{**}p < 0.01$, $^{***}p < 0.001$; ns = not significant.

| Comparison | Zone | DM stat | $p$-value | Sig. | Interpretation |
| --- | --- | --- | --- | --- | --- |
| LightGBM vs. Ridge ARX | NO1 | $-2.84$ | 0.002 | ** | LightGBM significantly better |
| | NO2 | $-4.14$ | <0.001 | *** | LightGBM significantly better |
| | NO3 | $-0.88$ | 0.190 | ns | Not significant |
| | NO4 | $+0.45$ | 0.672 | ns | Not significant |
| | NO5 | $-2.12$ | 0.017 | * | LightGBM significantly better |
| LightGBM vs. Naïve-24h | NO1 | $-9.14$ | <0.001 | *** | LightGBM significantly better |
| | NO2 | $-8.81$ | <0.001 | *** | LightGBM significantly better |
| | NO3 | $-6.18$ | <0.001 | *** | LightGBM significantly better |
| | NO4 | $-3.81$ | <0.001 | *** | LightGBM significantly better |
| | NO5 | $-8.68$ | <0.001 | *** | LightGBM significantly better |
| Ridge ARX vs. Naïve-24h | NO1 | $-8.95$ | <0.001 | *** | Ridge ARX significantly better |
| | NO2 | $-8.61$ | <0.001 | *** | Ridge ARX significantly better |
| | NO3 | $-6.17$ | <0.001 | *** | Ridge ARX significantly better |
| | NO4 | $-3.78$ | <0.001 | *** | Ridge ARX significantly better |
| | NO5 | $-8.55$ | <0.001 | *** | Ridge ARX significantly better |

Table S7: Leave-one-group-out (LOGO) feature ablation results across the five Norwegian bidding zones. For each feature group, the model was retrained after removing that group from the full specification. Positive $\Delta$MAE and $\Delta R^2$ values indicate that the removed group contributed positively to predictive performance in the full model.

| Zone | Feature group | MAE | $R^2$ | $\Delta$MAE | $\Delta R^2$ |
| --- | --- | --- | --- | --- | --- |
| NO1 | lags | 23.300 | 0.079 | 17.562 | 0.809 |
| NO1 | calendar | 6.097 | 0.878 | 0.359 | 0.010 |
| NO1 | weather | 5.733 | 0.891 | -0.005 | -0.003 |
| NO1 | reservoir | 5.663 | 0.890 | -0.075 | -0.002 |
| NO1 | commodities | 5.649 | 0.889 | -0.089 | -0.002 |
| NO1 | load_wsf | 5.786 | 0.886 | 0.048 | 0.001 |
| NO2 | lags | 19.554 | 0.235 | 14.231 | 0.657 |
| NO2 | calendar | 5.861 | 0.880 | 0.538 | 0.012 |
| NO2 | weather | 5.361 | 0.893 | 0.038 | -0.001 |
| NO2 | reservoir | 5.351 | 0.892 | 0.028 | 0.001 |
| NO2 | commodities | 5.334 | 0.893 | 0.011 | -0.001 |
| NO2 | load_wsf | 5.324 | 0.892 | 0.001 | 0.001 |
| NO3 | lags | 20.174 | -0.045 | 17.054 | 0.977 |

*(Table S7 continued)*

| Zone | Feature group | MAE | $R^2$ | $\Delta$MAE | $\Delta R^2$ |
|---|---|---|---|---|---|
| NO3 | calendar | 3.165 | 0.928 | 0.045 | 0.004 |
| NO3 | weather | 3.054 | 0.934 | -0.066 | -0.002 |
| NO3 | reservoir | 3.072 | 0.934 | -0.048 | -0.002 |
| NO3 | commodities | 3.052 | 0.932 | -0.068 | -0.000 |
| NO3 | load_wsf | 3.079 | 0.933 | -0.041 | -0.002 |
| NO4 | lags | 22.529 | -0.865 | 20.889 | 1.798 |
| NO4 | calendar | 1.702 | 0.929 | 0.062 | 0.003 |
| NO4 | weather | 1.654 | 0.936 | 0.014 | -0.004 |
| NO4 | reservoir | 1.633 | 0.934 | -0.007 | -0.002 |
| NO4 | commodities | 1.615 | 0.935 | -0.025 | -0.003 |
| NO4 | load_wsf | 1.635 | 0.936 | -0.005 | -0.004 |
| NO5 | lags | 26.716 | -0.372 | 22.879 | 1.283 |
| NO5 | calendar | 4.189 | 0.903 | 0.352 | 0.007 |
| NO5 | weather | 3.907 | 0.910 | 0.070 | 0.000 |
| NO5 | reservoir | 3.856 | 0.911 | 0.019 | -0.001 |
| NO5 | commodities | 3.855 | 0.910 | 0.018 | 0.001 |
| NO5 | load_wsf | 3.960 | 0.910 | 0.123 | 0.001 |

Table S8: Lags-plus-one-group ablation results across the five Norwegian bidding zones. Each model includes lagged prices and one additional feature group only. Negative $\Delta$MAE$_{\text{vs full}}$ and values of $\Delta R^2_{\text{vs full}}$ closer to zero indicate that the reduced model more closely matches or slightly exceeds the full model.

| Zone | Feature group | MAE | $R^2$ | $\Delta$MAE$_{\text{vs full}}$ | $\Delta R^2_{\text{vs full}}$ |
|---|---|---|---|---|---|
| NO1 | lags (only) | 5.808 | 0.882 | 0.070 | -0.006 |
| NO1 | calendar | 5.572 | 0.894 | -0.166 | 0.006 |
| NO1 | weather | 5.846 | 0.880 | 0.108 | -0.007 |
| NO1 | reservoir | 5.919 | 0.878 | 0.181 | -0.009 |
| NO1 | commodities | 5.946 | 0.879 | 0.208 | -0.008 |
| NO1 | load_wsf | 5.767 | 0.883 | 0.029 | -0.005 |
| NO2 | lags (only) | 5.544 | 0.883 | 0.221 | -0.009 |
| NO2 | calendar | 5.273 | 0.895 | -0.050 | 0.003 |
| NO2 | weather | 5.593 | 0.882 | 0.270 | -0.011 |
| NO2 | reservoir | 5.628 | 0.880 | 0.305 | -0.012 |
| NO2 | commodities | 5.591 | 0.882 | 0.268 | -0.010 |
| NO2 | load_wsf | 5.554 | 0.883 | 0.231 | -0.009 |
| NO3 | lags (only) | 3.077 | 0.929 | -0.043 | -0.003 |
| NO3 | calendar | 3.001 | 0.934 | -0.119 | 0.002 |

*(Table S8 continued)*

| Zone | Feature group | MAE | $R^2$ | $\Delta\text{MAE}_{\text{vs full}}$ | $\Delta R^2_{\text{vs full}}$ |
|---|---|---|---|---|---|
| NO3 | weather | 3.084 | 0.929 | -0.036 | -0.003 |
| NO3 | reservoir | 3.129 | 0.926 | 0.009 | -0.005 |
| NO3 | commodities | 3.097 | 0.928 | -0.023 | -0.003 |
| NO3 | load_wsf | 3.069 | 0.930 | -0.051 | -0.002 |
| NO4 | lags (only) | 1.644 | 0.929 | 0.004 | -0.004 |
| NO4 | calendar | 1.576 | 0.935 | -0.064 | 0.002 |
| NO4 | weather | 1.604 | 0.931 | -0.036 | -0.001 |
| NO4 | reservoir | 1.656 | 0.929 | 0.016 | -0.003 |
| NO4 | commodities | 1.658 | 0.930 | 0.018 | -0.002 |
| NO4 | load_wsf | 1.624 | 0.930 | -0.016 | -0.002 |
| NO5 | lags (only) | 3.941 | 0.906 | 0.104 | -0.005 |
| NO5 | calendar | 3.758 | 0.914 | -0.079 | 0.003 |
| NO5 | weather | 4.005 | 0.903 | 0.168 | -0.007 |
| NO5 | reservoir | 3.990 | 0.904 | 0.153 | -0.006 |
| NO5 | commodities | 4.004 | 0.904 | 0.167 | -0.007 |
| NO5 | load_wsf | 3.946 | 0.907 | 0.109 | -0.004 |

Table S9: Conditional regime analysis across all five Norwegian bidding zones for the reduced Lags+Calendar model and the full LightGBM model. Each row reports the number of hourly observations ($n$), mean and standard deviation of the observed price within the regime cell, and model performance measured by sMAPE and RMSE. Regime cells are defined by median splits of reservoir anomaly and TTF gas price within the 2025 test set.

| Zone | Model | Regime cell | $n$ | Mean price | Price SD | sMAPE (%) | RMSE |
|---|---|---|---|---|---|---|---|
| NO1 | Lags+Calendar | High res / High TTF | 3071 | 62.572 | 37.657 | 9.762 | 12.546 |
| NO1 | Full LightGBM | High res / High TTF | 3071 | 62.572 | 37.657 | 10.287 | 13.012 |
| NO1 | Lags+Calendar | High res / Low TTF | 1464 | 69.062 | 33.198 | 7.802 | 9.000 |
| NO1 | Full LightGBM | High res / Low TTF | 1464 | 69.062 | 33.198 | 8.205 | 9.495 |
| NO1 | Lags+Calendar | Low res / High TTF | 1320 | 47.396 | 23.107 | 24.916 | 10.061 |
| NO1 | Full LightGBM | Low res / High TTF | 1320 | 47.396 | 23.107 | 26.497 | 10.165 |
| NO1 | Lags+Calendar | Low res / Low TTF | 2881 | 53.234 | 27.600 | 19.379 | 9.415 |
| NO1 | Full LightGBM | Low res / Low TTF | 2881 | 53.234 | 27.600 | 20.103 | 9.534 |
| NO2 | Lags+Calendar | High res / High TTF | 3791 | 64.165 | 36.754 | 10.744 | 12.322 |
| NO2 | Full LightGBM | High res / High TTF | 3791 | 64.165 | 36.754 | 10.884 | 12.476 |
| NO2 | Lags+Calendar | High res / Low TTF | 744 | 58.029 | 22.264 | 14.010 | 7.709 |
| NO2 | Full LightGBM | High res / Low TTF | 744 | 58.029 | 22.264 | 13.823 | 7.816 |
| NO2 | Lags+Calendar | Low res / High TTF | 600 | 63.718 | 23.511 | 17.513 | 8.342 |
| NO2 | Full LightGBM | Low res / High TTF | 600 | 63.718 | 23.511 | 16.995 | 8.166 |
| NO2 | Lags+Calendar | Low res / Low TTF | 3601 | 68.371 | 29.325 | 12.284 | 8.860 |

*(Table S9 continued)*

| Zone | Model | Regime cell | $n$ | Mean price | Price SD | sMAPE (%) | RMSE |
|---|---|---|---|---|---|---|---|
| NO2 | Full LightGBM | Low res / Low TTF | 3601 | 68.371 | 29.325 | 12.063 | 9.061 |
| NO3 | Lags+Calendar | High res / High TTF | 3719 | 18.130 | 23.551 | 25.854 | 7.334 |
| NO3 | Full LightGBM | High res / High TTF | 3719 | 18.130 | 23.551 | 25.948 | 7.340 |
| NO3 | Lags+Calendar | High res / Low TTF | 744 | 16.875 | 18.365 | 29.819 | 4.187 |
| NO3 | Full LightGBM | High res / Low TTF | 744 | 16.875 | 18.365 | 30.683 | 4.239 |
| NO3 | Lags+Calendar | Low res / High TTF | 672 | 10.390 | 10.395 | 35.876 | 4.449 |
| NO3 | Full LightGBM | Low res / High TTF | 672 | 10.390 | 10.395 | 36.001 | 4.462 |
| NO3 | Lags+Calendar | Low res / Low TTF | 3601 | 26.798 | 25.406 | 17.410 | 5.326 |
| NO3 | Full LightGBM | Low res / Low TTF | 3601 | 26.798 | 25.406 | 17.893 | 5.523 |
| NO4 | Lags+Calendar | High res / High TTF | 3431 | 5.725 | 13.530 | 28.289 | 5.513 |
| NO4 | Full LightGBM | High res / High TTF | 3431 | 5.725 | 13.530 | 28.633 | 5.606 |
| NO4 | Lags+Calendar | High res / Low TTF | 1008 | 3.119 | 4.931 | 27.470 | 1.741 |
| NO4 | Full LightGBM | High res / Low TTF | 1008 | 3.119 | 4.931 | 27.895 | 1.732 |
| NO4 | Lags+Calendar | Low res / High TTF | 960 | 5.430 | 5.988 | 25.582 | 2.497 |
| NO4 | Full LightGBM | Low res / High TTF | 960 | 5.430 | 5.988 | 25.213 | 2.542 |
| NO4 | Lags+Calendar | Low res / Low TTF | 3337 | 14.069 | 25.121 | 20.621 | 4.886 |
| NO4 | Full LightGBM | Low res / Low TTF | 3337 | 14.069 | 25.121 | 21.185 | 4.979 |
| NO5 | Lags+Calendar | High res / High TTF | 3527 | 49.158 | 28.298 | 9.233 | 10.548 |
| NO5 | Full LightGBM | High res / High TTF | 3527 | 49.158 | 28.298 | 9.533 | 10.694 |
| NO5 | Lags+Calendar | High res / Low TTF | 1008 | 46.351 | 18.633 | 9.086 | 5.436 |
| NO5 | Full LightGBM | High res / Low TTF | 1008 | 46.351 | 18.633 | 9.555 | 5.581 |
| NO5 | Lags+Calendar | Low res / High TTF | 864 | 28.917 | 17.070 | 18.208 | 4.830 |
| NO5 | Full LightGBM | Low res / High TTF | 864 | 28.917 | 17.070 | 18.219 | 4.623 |
| NO5 | Lags+Calendar | Low res / Low TTF | 3337 | 49.203 | 31.563 | 13.686 | 7.064 |
| NO5 | Full LightGBM | Low res / Low TTF | 3337 | 49.203 | 31.563 | 13.611 | 7.308 |



\end{document}